\documentclass{article}

\usepackage{arxiv}

\usepackage{latexsym}
\usepackage[T1]{fontenc}
\usepackage[utf8]{inputenc}
\usepackage{microtype}
\usepackage{amsmath,amssymb}
\allowdisplaybreaks
\usepackage{booktabs}
\usepackage{graphicx}
\usepackage{xcolor}
\usepackage{caption}
\usepackage[round]{natbib}
\definecolor{divrblue}{HTML}{1A4C9C}
\usepackage[breaklinks,colorlinks=true,citecolor=divrblue,linkcolor=divrblue,urlcolor=divrblue]{hyperref}
\usepackage{listings}
\lstdefinestyle{prompt}{basicstyle=\ttfamily\scriptsize,breaklines=true,breakatwhitespace=false,columns=fullflexible,keepspaces=true,frame=single,framesep=2pt,xleftmargin=2pt,xrightmargin=2pt,aboveskip=4pt,belowskip=6pt,postbreak=\mbox{$\hookrightarrow$\space}}

\title{Learning When Diffusion Language Models Should Regenerate}
\author{%
\textbf{Minhae Oh}$^{1}$, \textbf{Jeonghye Kim}$^{2}$, \textbf{Jinwoo Koo}$^{1}$, \textbf{Yeongoon Kim}$^{1}$,\\
\textbf{Jimin Seo}$^{1}$, \textbf{Minsik Seol}$^{1}$, \textbf{Jungwoo Lee}$^{1,\dagger}$\\[3pt]
$^{1}$Seoul National University \quad $^{2}$KAIST\\[2pt]
\texttt{minhae.oh@cml.snu.ac.kr}, \texttt{junglee@snu.ac.kr}\\[2pt]
{\footnotesize $^{\dagger}$Corresponding author.}
}

\date{}
\renewcommand{\shorttitle}{Learning When Diffusion Language Models Should Regenerate}

\hypersetup{
pdftitle={Learning When Diffusion Language Models Should Regenerate},
pdfauthor={Minhae Oh, Jeonghye Kim, Jinwoo Koo, Yeongoon Kim, Jimin Seo, Minsik Seol, Jungwoo Lee},
}

\begin{document}
\raggedbottom
\maketitle
\begin{abstract}
Remasking allows diffusion language models to revisit generated text, but effective correction requires knowing when and what to revise. Our diagnosis on Sudoku reveals two gaps. Hidden states contain error information that token confidence does not reliably expose, and remasking an isolated error usually reproduces it when the surrounding context remains fixed. These findings motivate DiVR , a verifier that predicts whether a partially generated sequence will reach a correct answer. Trained on the final correctness of the model's own trajectories, DiVR decides when to regenerate a block and which alternative to retain, while the generator itself stays unchanged. On five reasoning benchmarks with LLaDA-8B-Base, DiVR raises the average accuracy of the fine-tuned generator from 35.3\% to 43.4\% at $1.9$--$3.5\times$ the cost of greedy decoding, above search baselines that use $3.8$--$7.4\times$. It also complements post-training such as supervised fine-tuning and reinforcement learning, and generalizes to other generators, including Dream.
\end{abstract}

\section{Introduction}
\label{sec:introduction}
Masked diffusion language models (MDLMs) generate text by repeatedly filling in the masked positions of a partially completed canvas, predicting every position with the whole response in view \citep{sahoo2024simple,nie2025large,ye2025dream}. This joint generation helps on planning tasks such as Sudoku and Countdown \citep{ye2025beyond}, reduces the number of sequential steps \citep{khanna2025mercury,wu2026fast}, and in principle allows generated tokens to be masked again and revised \citep{wang2025remasking}.

In practice, standard MDLM decoding never revisits a committed token \citep{nie2025large}, so a wrong answer committed early with high confidence tends to remain \citep{zhao2026lockr}. Remasking samplers and search methods add revision, but decide what to revise by a schedule, the model's confidence or likelihood, a planner or prompted self-verification \citep{wang2025remasking,hong2026wide,peng2025path,luo2026self,bai2026prism}, none of which is trained to predict whether the final answer is correct. Learned alternatives act on individual tokens, through scorers trained against a reference sequence, unmasking policies or uniform-noise training \citep{kim2025fine,huang2026remedi,zhang2025corrective,jazbec2026learning,vonrutte2025gidd}, although in free-form reasoning a token that departs from the reference may still belong to a correct solution. Others borrow an external process reward model trained on autoregressive solutions \citep{mounier2025r3,dang2026inference,misaki2026unmaskfork,miles2026stitching}, a second large model that never sees the generator's states.

Recent work shows that MDLM hidden states carry correctness information about the sequence as a whole. The final answer is often identifiable before decoding ends \citep{li2026diffusion}, correct intermediate answers are sometimes overwritten \citep{wang2026time}, and hidden states predict whether the output will be correct \citep{chiu2026probing,zhao2026lockr}. It remains open whether this information pinpoints which tokens are wrong, and whether locating an error is enough to repair it.

We answer both questions on Sudoku, where the unique solution labels every cell, and find two gaps. Error information is present at the level of individual cells, but confidence does not expose it, as a linear probe on hidden states separates wrong cells from correct ones better than confidence does (AUROC 0.838 versus 0.685). Locating an error, in turn, does not repair it. Re-predicting exactly the wrong cells fixes only 6 of 77 failures because the retained context keeps supporting the mistake, and reopening part of the preceding reasoning raises this only to 13. Repair therefore requires revising the context together with the error.

Guided by this diagnosis, we propose DiVR (\textbf{Di}ffusion \textbf{V}erifier for \textbf{R}egeneration).\footnote{Code: \url{https://github.com/minhaeoh/DiVR}} DiVR shares the generator's backbone, scores the partially masked canvas after each block, and regenerates the whole block when the score is low, keeping the best-scoring alternative or the original. It is trained only on the generator's own decodes labelled by final-answer correctness, so it needs answer checks but no token-level labels \citep{cobbe2021gsm8k,yu2023outcomesupervised}. Our contributions are as follows.
\begin{itemize}
\item A diagnosis on Sudoku showing that MDLM errors are readable from hidden states but persist when only the erroneous tokens are re-predicted (Section~\ref{sec:diagnosis}).
\item DiVR, an outcome-supervised verifier that shares the generator's weights and decides when to regenerate a block and which alternative to keep (Section~\ref{sec:method}).
\item Experiments on five math and code benchmarks and three generators, in which DiVR outperforms the search baselines SR-SMC and Prism at a fraction of their cost, transfers to benchmarks and generators it was not trained on, and adds to supervised and reinforcement-learning post-training. Most of its repairs begin in the first block of a solution, as the diagnosis predicts (Section~\ref{sec:experiments}).
\end{itemize}

\section{Diagnosis: Readable Errors, Persistent Mistakes}

We ask whether an MDLM's errors are represented in its hidden states, whether its confidence exposes them, and whether reopening an erroneous position repairs it. Sudoku answers these questions with token-level ground truth, because every cell of a uniquely solvable puzzle has exactly one correct digit.

\paragraph{Setup.}
We study LLaDA-8B-Instruct after one epoch of supervised training on Sudoku propagation traces, on the 190 uniquely solvable $4{\times}4$ evaluation puzzles of d1 \citep{zhao2025d1}, with a 128-token canvas, 128 steps and temperature zero. We analyze two decoders. \emph{Commit-once} fills masked positions in confidence order and never changes them, and \emph{confidence-based remasking} may mask filled positions again, so that the model can revise a committed mistake. Linear probes on answer-position hidden states are compared with confidence, scored as $1-p_{\mathrm{top1}}$, on the same observations (Appendix~\ref{sec:app-diagnosis}).

\begin{figure}[t]
    \centering
    \includegraphics[width=\textwidth]{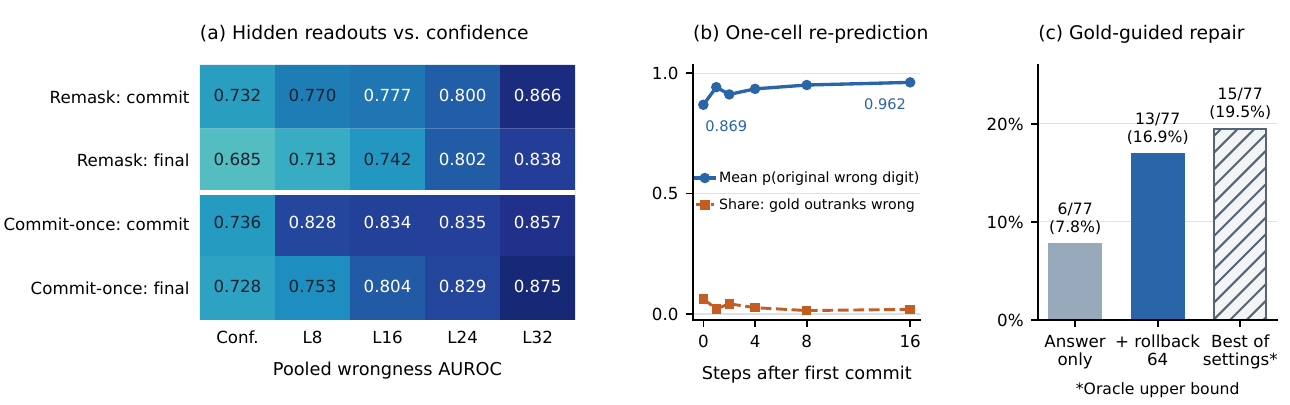}
    \caption{Diagnosis on uniquely solvable $4\times4$ Sudoku. \textbf{(a)} Wrongness AUROC of hidden-state probes and confidence on matched observations. \textbf{(b)} After re-masking a wrong cell, the original digit stays highly probable and the gold digit rarely outranks it. \textbf{(c)} Gold-guided repair on 77 failures: reopening the wrong cells alone fixes few, and the per-puzzle best setting is an oracle upper bound.}
    \label{fig:sudoku4-diagnosis}
\end{figure}

\paragraph{Finding 1. Representations expose errors more reliably than confidence.}
On remasking trajectories, the final-layer probe separates wrong from correct cells with AUROC 0.838 at the final state, against 0.685 for confidence (Figure~\ref{fig:sudoku4-diagnosis}a). The information is already present when a digit is first chosen. At the forward pass that first fills a cell, while the cell is still masked, the probe reaches 0.866 against 0.732. Commit-once trajectories show the same ordering, and the probe also ranks the wrong cells of a puzzle above its correct ones, so it locates errors rather than only recognizing difficult puzzles.

\label{sec:diagnosis}

\paragraph{Finding 2. Locating an error does not guarantee repair under its context.}
We mask one committed wrong cell, keep the rest of the canvas, and let the model predict it again. The original wrong digit keeps a mean probability of 0.869, rising to 0.962 when the cell is masked sixteen steps after its commit, and the gold digit rarely outranks it (Figure~\ref{fig:sudoku4-diagnosis}b). On 77 failed puzzles, a gold oracle that reopens every wrong answer cell repairs only 6 (Figure~\ref{fig:sudoku4-diagnosis}c). Reopening the reasoning written after the first error as well helps, but the best fixed rollback depth repairs 13, and even the best depth chosen for each puzzle with the gold answer repairs 15.

\paragraph{Implication.}
Error information can be read from the model's states, but re-predicting the erroneous tokens under their context rarely repairs them. Instead of locating an error and fixing it in place, a method should regenerate a region jointly and judge the alternatives by their eventual outcome, which is what DiVR does.

\section{Method}
\label{sec:method}
\subsection{Learning an outcome criterion on masked states}
\label{sec:method-states}

\begin{figure}[htb]
    \centering
    \includegraphics[width=\textwidth]{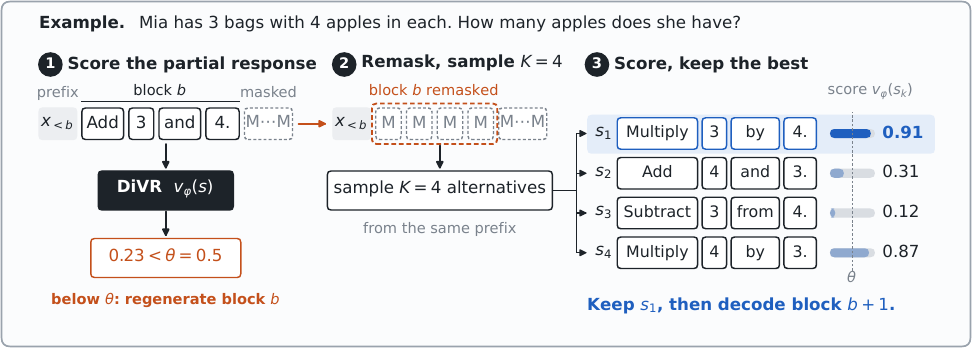}
    \caption{\textbf{DiVR-guided block re-decoding.} DiVR scores the canvas, masked suffix included, after each block ($B=4$ here, 32 in the experiments). A score below $\theta$ triggers four alternatives of the block from the same prefix, and the best-scoring state ($s_1$, 0.91) replaces the original. Text and scores are illustrative.}
    \label{fig:vrm-inference}
\end{figure}

The diagnosis motivates two design choices. The criterion for revision should be learned rather than read from token confidence, and the unit of revision should be a region regenerated jointly rather than an isolated token. DiVR implements both with an outcome-supervised verifier and block re-decoding, as Figure~\ref{fig:vrm-inference} illustrates.

Let $q$ be a prompt, $L$ the response-canvas length, and $B$ the block length. For a sampled response $y_i$, the state after block $b$ (numbered from one) is
\begin{equation}
    s_{i,b} = [q;\, y_{i,1:bB};\, \mathrm{MASK}^{L-bB}].
    \label{eq:vrm-state}
\end{equation}
We label every block-boundary state of a sampled response with the binary outcome $r_i$ of the completed response. Following outcome-supervised verification \citep{cobbe2021gsm8k}, this needs only an answer check on the finished response, not gold reasoning steps or token-level labels. A negative label says that the trajectory through a state failed, not that the state itself is wrong. The training states are block prefixes with masked suffixes, exactly the states that decoding visits, and they come from the generator's own samples, so the verifier learns the error modes of the generator it guides.

\subsection{A verifier that shares the generator's weights}
\label{sec:method-head}

The verifier runs the generator's backbone on a state and maps the pooled final-layer states to a score with a small head. Its own parameters $\phi$ are this head and a low-rank adapter (LoRA) on the backbone's attention and feed-forward projections \citep{hu2022lora}, trained together as one model, while the backbone weights stay frozen and shared with the generator. For a state $s$, the verifier computes
\begin{equation}
\begin{aligned}
    h_\phi(s) &= f_{W+\Delta W_\phi}(s), \\
    z_\phi(s) &= \operatorname{MLP}_\phi\!\left(
        \frac{1}{L}\sum_{t=1}^{L}h_\phi(s)_t\right), \\
    v_\phi(s) &= \sigma(z_\phi(s)),
\end{aligned}
    \label{eq:vrm-score}
\end{equation}
where $f_W$ is the generator's backbone with frozen weights $W$, $\Delta W_\phi$ is the adapter's low-rank update, and $h_\phi(s)_t$ is the final-layer state at response position $t$. Pooling averages all $L$ response positions, including those still masked, and excludes the prompt. The head $\operatorname{MLP}_\phi$ is a two-layer perceptron, and $\sigma$ is the logistic function. Training the adapter and the head together on the outcome objective lets the verifier reshape the backbone's representation of a state so that its information about the outcome is easier to use. The verifier's trainable parameters amount to about 1\% of the backbone.

\subsection{Classification and within-question ranking}
\label{sec:method-loss}

Each training step draws one question and one block index $b$ and forms the state $s_{i,b}$ of each of its four sampled trajectories, using the final boundary of a trajectory that ended earlier. The loss has one term for each decision of Section~\ref{sec:method-decoding}. A class-balanced binary cross-entropy $\mathcal{L}_{\mathrm{cls}}$ trains each score to predict its trajectory's outcome, and a pairwise logistic loss $\mathcal{L}_{\mathrm{rank}}$ trains correct trajectories to score above incorrect ones. With $z_i=z_\phi(s_{i,b})$, we minimize
\begin{align}
    \mathcal{L}_{\mathrm{cls}} &= \frac{1}{4}\sum_i
        w_{r_i}\,\ell_{\mathrm{BCE}}(z_i,r_i), \\
    \mathcal{L}_{\mathrm{rank}} &= \frac{1}{|\mathcal{P}|}
        \sum_{(i,j)\in\mathcal{P}}\log(1+e^{z_j-z_i}), \\
    \mathcal{L} &= \mathcal{L}_{\mathrm{cls}}+\mathcal{L}_{\mathrm{rank}},
    \label{eq:vrm-loss}
\end{align}
where $w_1=1/(2\pi)$ and $w_0=1/[2(1-\pi)]$ give correct and incorrect trajectories equal total weight, $\pi$ is the fraction of correct trajectories in the sampled pool, and $\mathcal{P}$ contains the correct--incorrect pairs of the question. The classification term places the scores on an absolute scale, on which $\theta=1/2$ asks whether a state is less promising than an average sampled trajectory. The ranking term trains the within-question ordering that selecting among alternatives requires.

\subsection{Selection and block re-decoding}
\label{sec:method-decoding}

For best-of-$k$, the generator samples $k$ complete responses, and we return the one whose final state scores highest. For guided decoding, the generator fills each block in its ordinary way and the verifier scores the canvas. If the score is below $\theta$ and the regeneration budget is not exhausted, we remask the whole block, sample $K$ alternatives from the same prefix, and keep the original unless an alternative scores higher. Decoding then continues, and every regeneration event counts against the budget. Because the whole block is resampled, its tokens change together, as the diagnosis requires, and the procedure needs neither to locate the error nor to know what caused it.

\section{Experiments}
\label{sec:experiments}
\begin{table}[t]
\caption{\textbf{Main results across three generators.} Each panel is one model and each block one generator (untuned or SFT for LLaDA-8B-Base, released or diffu-GRPO for LLaDA-8B-Instruct); every method in a block runs on that generator, with one verifier for the math columns and one for the code columns. Cost is forwards relative to the generator's paired greedy run, verification included. \textbf{Bold} marks the best decoding method of a generator; the best-of-4 rows use about four times the greedy budget and are not ranked. $^\ast$Verified by the verifiers of the SFT generator. $^\ddagger$After 3{,}800 steps of d1's diffu-GRPO recipe on GSM8K \citep{zhao2025d1}, with verifiers trained on its own decodes (Table~\ref{tab:rl}). }
\label{tab:main-results}
\centering
\small
\setlength{\tabcolsep}{2.5pt}
\renewcommand{\arraystretch}{1.08}
\definecolor{divrblue}{HTML}{1A4C9C}
\newcommand{\mainours}[1]{\textcolor{divrblue}{\textbf{#1}}}
\begin{tabular*}{\textwidth}{@{\extracolsep{\fill}}l*{5}{r@{\hspace{4pt}}l}r@{}}
\toprule
\textbf{Model / method} & \multicolumn{6}{c}{\textbf{Math}} & \multicolumn{4}{c}{\textbf{Code}} & \\
\cmidrule(lr){2-7}\cmidrule(l){8-11}
& \multicolumn{2}{c}{GSM8K} & \multicolumn{2}{c}{MATH-500} & \multicolumn{2}{c}{Countdown} & \multicolumn{2}{c}{HumanEval} & \multicolumn{2}{c}{MBPP} & \multicolumn{1}{c}{Avg.} \\
\cmidrule(lr){2-3}\cmidrule(lr){4-5}\cmidrule(lr){6-7}\cmidrule(lr){8-9}\cmidrule(lr){10-11}\cmidrule(l){12-12}
& \multicolumn{1}{c}{Acc.} & Cost & \multicolumn{1}{c}{Acc.} & Cost & \multicolumn{1}{c}{Acc.} & Cost & \multicolumn{1}{c}{Acc.} & Cost & \multicolumn{1}{c}{Acc.} & Cost & \multicolumn{1}{c}{Acc.} \\
\midrule\midrule
\multicolumn{12}{@{}l}{\textbf{LLaDA-8B-Base}} \\
\quad Greedy & 61.8 & $\times1.0$ & 24.2 & $\times1.0$ & 10.6 & $\times1.0$ & 19.5 & $\times1.0$ & 30.2 & $\times1.0$ & 29.3 \\
\quad \mainours{+ DiVR-guided}$^\ast$ & \textbf{71.7} & $\times1.9$ & \textbf{26.8} & $\times1.7$ & \textbf{30.2} & $\times2.6$ & \textbf{23.8} & $\times2.7$ & \textbf{37.6} & $\times2.8$ & \textbf{38.0} \\
\midrule
\quad + SFT & 68.7 & $\times1.0$ & 24.6 & $\times1.0$ & 35.6 & $\times1.0$ & 23.2 & $\times1.0$ & 24.2 & $\times1.0$ & 35.3 \\
\quad + SFT + SR-SMC (4) & 72.6 & $\times4.0$ & 28.4 & $\times3.9$ & 43.0 & $\times3.9$ & \textbf{25.0} & $\times4.0$ & 26.6 & $\times3.8$ & 39.1 \\
\quad + SFT + Prism & 71.9 & $\times5.9$ & 32.4 & $\times6.0$ & 39.8 & $\times7.0$ & 20.1 & $\times5.2$ & 22.4 & $\times7.4$ & 37.3 \\
\quad \mainours{+ SFT + DiVR-guided} & \textbf{78.8} & $\times2.1$ & \textbf{33.8} & $\times1.9$ & \textbf{47.2} & $\times3.5$ & 23.8 & $\times2.9$ & \textbf{33.6} & $\times3.3$ & \textbf{43.4} \\
\quad \mainours{+ SFT + DiVR best-of-4} & 82.2 & $\times4.0$ & 34.2 & $\times3.7$ & 49.2 & $\times4.0$ & 26.2 & $\times4.0$ & 32.6 & $\times3.7$ & 44.9 \\
\midrule\midrule
\multicolumn{12}{@{}l}{\textbf{LLaDA-8B-Instruct}} \\
\quad Greedy & 79.2 & $\times1.0$ & 35.0 & $\times1.0$ & 15.6 & $\times1.0$ & 34.1 & $\times1.0$ & 33.2 & $\times1.0$ & 39.4 \\
\quad + SR-SMC (4) & 81.5 & $\times4.1$ & 41.2 & $\times4.0$ & 20.0 & $\times3.8$ & 32.3 & $\times4.0$ & 34.0 & $\times4.1$ & 41.8 \\
\quad + Prism & \textbf{83.5} & $\times5.0$ & \textbf{43.4} & $\times4.8$ & 21.2 & $\times4.9$ & 28.0 & $\times5.1$ & 31.6 & $\times5.6$ & 41.5 \\
\quad \mainours{+ DiVR-guided} & 83.0 & $\times1.6$ & 38.2 & $\times1.8$ & \textbf{23.0} & $\times2.6$ & \textbf{39.0} & $\times2.6$ & \textbf{40.8} & $\times2.5$ & \textbf{44.8} \\
\quad \mainours{+ DiVR best-of-4} & 87.6 & $\times4.0$ & 45.6 & $\times3.9$ & 35.4 & $\times4.0$ & 40.2 & $\times4.0$ & 40.2 & $\times4.1$ & 49.8 \\
\midrule
\quad + diffu-GRPO$^\ddagger$ & 77.4 & $\times1.0$ & \textbf{39.4} & $\times1.0$ & 17.0 & $\times1.0$ & 31.1 & $\times1.0$ & 35.4 & $\times1.0$ & 40.1 \\
\quad \mainours{+ diffu-GRPO + DiVR-guided}$^\ddagger$ & \textbf{83.4} & $\times1.6$ & 39.2 & $\times1.8$ & \textbf{28.2} & $\times2.7$ & \textbf{34.8} & $\times2.7$ & \textbf{41.4} & $\times2.3$ & \textbf{45.4} \\
\quad \mainours{+ diffu-GRPO + DiVR best-of-4}$^\ddagger$ & 87.4 & $\times4.0$ & 44.8 & $\times3.9$ & 42.0 & $\times4.0$ & 36.0 & $\times4.0$ & 42.0 & $\times4.0$ & 50.4 \\
\midrule\midrule
\multicolumn{12}{@{}l}{\textbf{Dream-v0-Instruct-7B}} \\
\quad Greedy & 77.3 & $\times1.0$ & 40.8 & $\times1.0$ & 18.0 & $\times1.0$ & 45.1 & $\times1.0$ & 43.8 & $\times1.0$ & 45.0 \\
\quad + SR-SMC (4) & 81.5 & $\times3.9$ & 45.8 & $\times3.9$ & 23.6 & $\times3.6$ & \textbf{53.0} & $\times4.2$ & 50.2 & $\times4.2$ & 50.8 \\
\quad + Prism & \textbf{85.1} & $\times5.5$ & \textbf{48.2} & $\times5.4$ & 29.2 & $\times5.0$ & 42.7 & $\times5.9$ & 46.4 & $\times7.5$ & 50.3 \\
\quad \mainours{+ DiVR-guided} & 84.4 & $\times1.7$ & 44.4 & $\times1.9$ & \textbf{37.2} & $\times2.4$ & \textbf{53.0} & $\times2.0$ & \textbf{55.2} & $\times2.9$ & \textbf{54.8} \\
\quad \mainours{+ DiVR best-of-4} & 89.6 & $\times4.0$ & 50.8 & $\times3.9$ & 42.4 & $\times3.8$ & 57.3 & $\times4.0$ & 55.0 & $\times4.0$ & 59.0 \\
\bottomrule
\end{tabular*}
\end{table}
\subsection{Experimental setup and baselines}

\paragraph{Models and benchmarks.}
We evaluate three models, LLaDA-8B-Base and LLaDA-8B-Instruct \citep{nie2025large} and Dream-v0-Instruct-7B \citep{ye2025dream}. LLaDA-8B-Base is evaluated both untuned (Base) and after supervised fine-tuning on GSM8K (SFT). The instruction-tuned models are used as released, and Appendix Table~\ref{tab:inst-sft} reports them after the same fine-tuning. We use three math benchmarks, GSM8K (1,319 problems) \citep{cobbe2021gsm8k}, MATH-500 (500) \citep{hendrycks2021math,lightman2023lets} and Countdown (500 three-number tasks) \citep{pan2025tinyzero}, and two code benchmarks, HumanEval (164) \citep{chen2021codex} and MBPP (500) \citep{austin2021program}, and report their unweighted average. The prompts are listed in Appendix~\ref{sec:prompts}. Each generator has one verifier per domain, trained on four of its own responses, sampled at temperature 0.7, for each of 2,000 GSM8K training questions for math and for each MBPP training and validation problem for code. A verifier serves every benchmark of its domain, and the untuned Base model is guided by the SFT generator's verifiers. The epoch and the threshold $\theta$ are chosen on development data, and the remaining hyperparameters are given in Appendix~\ref{sec:app-verifier}.

\paragraph{Decoding and baselines.}
Canvases contain 256 tokens, or 512 for MATH-500, decoded in 32-token blocks with 16 steps per block. Guided decoding uses $\theta=0.5$ and $K=4$ alternatives at temperature 0.7, with at most three regeneration events per problem. Cost counts forward passes, verifier reads included, relative to a greedy decode of the same problems, which is also the reference for accuracy. We compare against two search methods for diffusion decoding, reimplemented on the same generators and decoder (Appendix~\ref{sec:app-baselines}). SR-SMC \citep{luo2026self} keeps four particles and resamples them after each block by their accumulated log-confidence, and Prism \citep{bai2026prism} starts sixteen trajectories, prunes them to four with the generator's own prompted Yes/No judgment, and answers by majority vote, as in its paper (Appendix Table~\ref{tab:prism-variants} reports its released top-score rule). Both search with the generator's own judgment and keep several trajectories for every problem, whereas guided decoding branches only at flagged blocks. DiVR best-of-4 reranks four complete samples by the verifier's score at about the baselines' budget, which isolates the verifier's selection quality from block regeneration.

\subsection{Main results}
Table~\ref{tab:main-results} summarizes the results. Across the three generators, guided decoding is the most accurate method on average for every generator, and with best-of-4, which uses about the search baselines' budget, DiVR is the most accurate on every math and code benchmark. On the SFT generator, guided decoding raises GSM8K from 68.7\% to 78.8\% and MBPP from 24.2\% to 33.6\%, and best-of-4 raises GSM8K further to 82.2\%. On GSM8K with the SFT generator, guided decoding is also more accurate than the search baselines at a lower cost, reaching 78.8\% at $2.1\times$ the cost of greedy decoding against 72.6\% at $4.0\times$ for SR-SMC and 71.9\% at $5.9\times$ for Prism. On LLaDA-8B-Instruct, guided decoding raises the five-benchmark average from 39.4\% to 44.8\%, and best-of-4 reaches 49.8\%, 10.4 points above greedy decoding. These results show that DiVR is an effective inference-time revision method for MDLMs.

\paragraph{DiVR is cost-effective.}
Over the five benchmarks of the SFT generator, guided decoding raises the average from 35.3\% to 43.4\% at $1.9$--$3.5\times$ the cost of greedy decoding, whereas SR-SMC reaches 39.1\% at $3.8$--$4.0\times$ and Prism 37.3\% at $5.2$--$7.4\times$. On average, guided decoding thus uses about 70\% of SR-SMC's computation and less than half of Prism's. Best-of-4 with the same verifier is the most accurate row on most benchmarks, and guided decoding comes close to it at as little as 40\% of its cost. Guided decoding retains most of the gain of best-of-4 because it samples alternatives only for the blocks the verifier flags, which Table~\ref{tab:regen-blocks} shows to be concentrated at the start of a solution. The gap between the two DiVR rows is small next to the gap between either of them and the baselines, so the benefit comes from the verifier's judgment rather than from the amount of sampling.

\paragraph{DiVR complements post-training.}
We test whether DiVR adds to post-training of the generator on two post-trained variants, supervised fine-tuning of LLaDA-8B-Base and diffu-GRPO reinforcement learning of LLaDA-8B-Instruct, the model for which d1 published its recipe \citep{zhao2025d1}, trained at about one eighth of d1's GSM8K budget. DiVR improves both. Fine-tuning raises Countdown from 10.6\% to 35.6\%, and guided decoding raises it further to 47.2\%. For the RL generator, guided decoding raises the five-benchmark average from 40.1\% to 45.4\%. RL training moved the Instruct run's reward, with its correctness term rising from 0.76 to 1.35 of a possible 2, but greedy accuracy moved little (the rows marked $\ddagger$). With guidance, the RL generator gains 6.0 points on GSM8K and 11.2 on Countdown over its own greedy decoding, more than the verifier adds to Instruct itself (3.8 and 7.4). RL changes what the generator writes, the verifier decides when a block is discarded, and the two gains add. Appendix Table~\ref{tab:rl} gives the training details.

\paragraph{DiVR generalizes across benchmarks and generators.}
Each verifier is trained on a single benchmark per domain, GSM8K for math and MBPP for code, yet it also improves benchmarks it never saw. The GSM8K-trained verifier raises the SFT generator on MATH-500 from 24.6\% to 33.8\% and on Countdown from 35.6\% to 47.2\%, and the MBPP-trained verifier raises HumanEval, which has no training split, for LLaDA-8B-Instruct and Dream (34.1\% to 39.0\% and 45.1\% to 53.0\%) (Appendix Table~\ref{tab:code-transfer} compares the math and code verifiers on code). The verifiers also transfer across generators. Without any training on the untuned Base model's decodes, the SFT generator's verifiers raise its average from 29.3\% to 38.0\%, above the 35.3\% of fine-tuning alone. Trained on its own decodes, the same recipe raises Dream-v0-Instruct-7B, a second model family with a different tokenizer, from 45.0\% to 54.8\%. The gains are largest where the generator's own signals are least reliable, as on Countdown, where Dream's accuracy doubles from 18.0\% to 37.2\%.

\paragraph{DiVR repairs errors early.}

\begin{table}[t]
\caption{\textbf{Where guided decoding repairs the greedy answer.} SFT generator on the math benchmarks. Fixed counts the greedy errors that guided decoding corrects, out of 413 on GSM8K, 377 on MATH-500 and 322 on Countdown, and the repair block is the first block at which an alternative replaced the original. TPR and FPR treat a wrong greedy answer as positive and a first-block score below $\theta$ as a flag. Most repairs begin in the first block.}
\label{tab:regen-blocks}
\centering
\small
\setlength{\tabcolsep}{7pt}
\begin{tabular}{lrrrrrrr}
\toprule
& & \multicolumn{4}{c}{Repair block} & \multicolumn{2}{c}{Flagged at block 0} \\
\cmidrule(lr){3-6}\cmidrule(l){7-8}
Benchmark & Fixed & 0 & 1 & 2 & $\geq$3 & TPR & FPR \\
\midrule
GSM8K & 154 & 114 & 28 & 7 & 5 & 76\% & 22\% \\
MATH-500 & 52 & 25 & 15 & 9 & 3 & 50\% & 14\% \\
Countdown & 59 & 58 & 1 & 0 & 0 & 99\% & 1\% \\
\bottomrule
\end{tabular}
\end{table}

To see where DiVR triggers the regenerations that repair an answer, we locate the repairs on the three math benchmarks. For every problem that greedy decoding gets wrong and guided decoding gets right, we record the block at which an alternative first replaced the original, which is where the guided trajectory departs from the greedy one. As Table~\ref{tab:regen-blocks} shows, the repairs are concentrated at the start of the solution. On GSM8K, 74\% of the 154 repaired problems depart in the first block and 92\% within the first two. On Countdown, 58 of the 59 repairs start in the first block, because the generator writes its whole candidate expression there and the next block only copies it into the answer. Appendix~\ref{sec:examples} shows three such repairs.

This concentration has two sources, and the table separates them. The first is detection. Among greedy decodes, the first-block score falls below $\theta$ for 76\% of the eventually wrong GSM8K solutions and 22\% of the eventually right ones, for 99\% and 1\% on Countdown, and for 50\% and 14\% on MATH-500. One block is therefore enough for the verifier to separate most trajectories by their eventual outcome. The second source is repairability, which the diagnosis predicted. The first block fixes the interpretation of the problem, such as which quantity is asked for or how the equation is set up, and later tokens are predicted from that context and tend to confirm it. A fresh sample of the first block starts from the prompt alone and can take a different route, whereas a fresh sample of a later block inherits the mistaken setup.

\section{Analysis}
\label{sec:analysis}

\begin{table}[ht]
\caption{\textbf{Same block re-decoding, different scores.} GSM8K test with the SFT generator (paired greedy 68.7). All signals share $K$, the temperature and the redo cap, and each threshold is the best on the development split. $C_{\mathrm{total}}$ is the verification-inclusive cost relative to greedy.}
\label{tab:signal-control}
\centering
\small
\setlength{\tabcolsep}{7pt}
\renewcommand{\arraystretch}{1.13}
\begin{tabular}{@{}lrr@{}}
\toprule
Evaluation signal & Success (\%) & $C_{\mathrm{total}}$ \\
\midrule
Commit confidence & 72.2 & $2.45\times$ \\
Prompted Yes/No & 67.9 & $3.62\times$ \\
Head-only verifier (V0) & 74.2 & $3.07\times$ \\
\midrule
\textbf{DiVR} & \textbf{78.8} & $2.10\times$ \\
\bottomrule
\end{tabular}
\end{table}

\paragraph{Controlling the evaluation signal.}
The main results combine a learned score with a regeneration procedure, so we first ask whether the improvement comes from the verifier or from the opportunity to regenerate. On GSM8K with the SFT generator, we hold the regeneration procedure and its budget fixed and replace only the signal that decides when to regenerate and which block to keep. As shown in Table~\ref{tab:signal-control}, three signals serve as controls. Commitment confidence is the geometric mean of the probabilities with which the visible tokens were placed, the signal on which confidence-based remasking relies. Prompted Yes/No asks the generator itself, through a judgment prompt, whether the partial solution is correct, the signal that Prism uses to prune its search. A head-only verifier (V0) is DiVR without its adapter, so its head is trained with the same outcome labels on the generator's unchanged hidden states. Each signal uses the threshold that maximizes its accuracy on the development split.

The ordering of the signals mirrors the diagnosis. Confidence and self-judgment reach 72.2\% and 67.9\%, so regenerating on the generator's own signals recovers a small part of the gain over the 68.7\% of greedy decoding or none of it. The head-only verifier reaches 74.2\%, so an outcome-trained score helps even on the generator's unchanged states. The full verifier reaches 78.8\% at the lowest cost, $2.10\times$ against $2.45$--$3.62\times$, which is consistent with a sharper signal spending fewer regeneration events on blocks that were already adequate. The improvement therefore comes from the verifier's learned criterion. The gap between the head-only verifier and DiVR also shows that reshaping the representation with the adapter, so that the outcome information is easier to use, matters.

\begin{table}[ht]
\caption{\textbf{Selection and ranking on a fixed candidate pool.} Success (\%) on the same four SFT-generator responses per problem. Oracle pass@4 is the pool's ceiling, and a uniformly chosen candidate scores 68.7, 24.6 and 33.7. Pair accuracy is the share of correct--incorrect candidate pairs of a question that DiVR ranks correctly from their states after each block.}
\label{tab:selection-analysis}
\centering
\small
\setlength{\tabcolsep}{7pt}
\renewcommand{\arraystretch}{1.08}
\begin{tabular}{@{}lrrr@{}}
\toprule
Selector & GSM8K & MATH-500 & Countdown \\
\midrule
Majority vote & 75.7 & 28.8 & 38.2 \\
Mean log-probability & 68.5 & 23.6 & 34.6 \\
\textbf{DiVR best-of-4} & \textbf{82.2} & \textbf{34.2} & \textbf{49.2} \\
\midrule
Oracle pass@4 & 85.4 & 43.4 & 49.8 \\
\midrule
\multicolumn{4}{@{}l}{\emph{DiVR pair accuracy}} \\
After block 1 & 0.705 & 0.665 & 0.957 \\
After block 2 & 0.830 & 0.735 & 0.971 \\
After block 3 & 0.879 & 0.760 & 0.971 \\
Finished & 0.919 & 0.796 & 0.971 \\
\bottomrule
\end{tabular}
\end{table}

\paragraph{Selection and ranking on fixed candidates.}
To separate the verifier's ranking quality from the regeneration procedure, Table~\ref{tab:selection-analysis} compares selectors on the same four sampled responses per problem, so that every selector chooses from an identical pool. DiVR exceeds majority voting on GSM8K (82.2\% against 75.7\%), MATH-500 (34.2\% against 28.8\%) and Countdown (49.2\% against 38.2\%). Mean log-probability reaches only 68.5\%, 23.6\% and 34.6\%, close to picking a candidate uniformly at random (68.7\%, 24.6\% and 33.7\%), so the generator's likelihood carries almost no information about which of its own samples is correct. On Countdown, where voting groups equivalent expressions, DiVR approaches the pool's oracle ceiling of 49.8\%, which means that it recognizes a correct expression whenever the pool contains one.

Guiding decoding requires a score that ranks partial solutions, not only finished ones. On the same candidates, DiVR's pair accuracy on GSM8K is already 0.705 after the first block and rises to 0.919 for the finished responses, and on Countdown it reaches 0.957 after the first block (Table~\ref{tab:selection-analysis}). Useful discrimination thus appears after the first block, where most repairs begin (Table~\ref{tab:regen-blocks}).

\paragraph{Sensitivity to the threshold.}
\begin{figure}[t]
    \centering
    \includegraphics[width=0.6\textwidth]{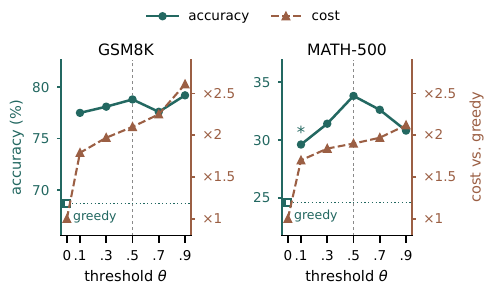}
    \caption{\textbf{Sensitivity to the regeneration threshold.} SFT generator, $K{=}4$, at most three regenerations, on the full test sets. Solid lines give success (\%) and dashed lines the verification-inclusive cost relative to greedy (square). The vertical line is $\theta=0.5$, chosen on the GSM8K development set beforehand, and $^\ast$ marks a paired McNemar difference from it at $p<0.05$.}
    \label{fig:threshold}
\end{figure}

The threshold $\theta=0.5$ was chosen on the GSM8K development split, and Figure~\ref{fig:threshold} varies it on the test sets with everything else fixed. On GSM8K, accuracy stays within 77.5--79.2\% for every $\theta$ from 0.1 to 0.9 while the cost rises from $1.79\times$ to $2.61\times$, so the threshold is primarily a cost control. On MATH-500, which lies outside the domain the threshold was chosen on, accuracy nonetheless peaks at the development-chosen $\theta=0.5$. A low threshold misses errors, and a high threshold regenerates correct work more often. The threshold chosen on in-domain development data is thus also the best setting out of domain. Appendix~\ref{sec:calibration} evaluates the scores as probabilities of success and the regenerate-or-continue decision at each threshold on paired states.

\section{Related Work}
\label{sec:related}
\paragraph{Refinement and search in MDLMs.}
MDLMs can revise intermediate states during iterative unmasking \citep{nie2025large,ye2025dream}. ReMDM remasks generated tokens to spend additional inference computation on refinement \citep{wang2025remasking}, and PRISM learns token-quality estimates to choose which positions to remask \citep{kim2025fine}. Search methods weight and resample trajectories by their accumulated log-probabilities \citep{luo2026self} or prune them with prompted self-verification \citep{bai2026prism}. Others bring an external process reward model into diffusion decoding to remask low-scoring blocks \citep{mounier2025r3}, resample particles \citep{dang2026inference}, guide a tree search over unmasking actions \citep{misaki2026unmaskfork} or stitch high-scoring steps across traces \citep{miles2026stitching}. Reinforcement learning with outcome or denoising-process rewards also improves reasoning in these models \citep{zhao2025d1,xie2026advancing}, and Section~\ref{sec:experiments} shows that DiVR's gains add to d1's recipe. DiVR differs from these methods in its signal, an outcome criterion learned on the generator's own partially masked states.

\paragraph{Error signals in MDLMs.}
In autoregressive LLMs, hidden states encode the correctness of outputs beyond what output probabilities reveal \citep{orgad2025llms}. For MDLMs, corrective post-training supervises incorrect tokens directly so that confidence-guided refinement targets them \citep{zhang2025corrective}, linear probes on hidden states predict whether the final output will be correct \citep{chiu2026probing}, and LOCKR detects answers that settle early on a wrong value from hidden-state trajectories and repairs them by remasking an answer region \citep{zhao2026lockr}. Concurrently, \citet{zhan2026why} train a verifier of a similar form, a LoRA-adapted MDLM backbone with an MLP head, and use it to prune candidates at every scoring step. Our work differs in three respects. Our diagnosis shows that a correctly located error is rarely repaired while its context is kept, which motivates regenerating whole blocks rather than answer tokens. Our verifier triggers regeneration only below a threshold and keeps the original block unless an alternative scores higher. Finally, we evaluate one procedure across three generators and five math and code benchmarks against confidence, self-verification and search baselines.

\paragraph{Outcome supervision and verifier-guided reasoning.}
\citet{cobbe2021gsm8k} train verifiers on sampled solutions labelled by final-answer correctness and use them to select among candidates, and process supervision labels individual steps instead \citep{lightman2023lets}. Outcome labels also support evaluating partial solutions, through values learned for step-level planning \citep{yu2023outcomesupervised} or process rewards derived from outcome supervision \citep{yuan2024implicitprm}, and verifiers can steer generation before a solution is complete \citep{liao2025reward}. For masked diffusion outside text, MDM-VGB regresses a verifier of masked states onto the terminal rewards of the model's own rollouts to guide unmasking and remasking \citep{jeon2026vgb}. DiVR brings such a state evaluator to partially masked text, trained on the generator's own decodes instead of a reward model trained on complete solutions, and uses it to decide block regeneration.

\section{Conclusion}
\label{sec:conclusion}
Our diagnosis shows that an MDLM can carry information about its errors and still fail to repair them in place, because the surrounding context keeps reproducing the mistake. DiVR therefore decides when a block should be written again, using a verifier learned from the generator's own outcomes. This single decision, made after each block, improves three generators on five math and code benchmarks, outperforms search at a fraction of its cost, and adds to supervised and reinforcement-learning post-training. Revision in MDLMs needs a learned criterion of when to start over more than a way to find the error.

\bibliography{references}

@inproceedings{nie2025large,
  title={Large Language Diffusion Models},
  author={Nie, Shen and Zhu, Fengqi and You, Zebin and Zhang, Xiaolu and Ou, Jingyang and Hu, Jun and Zhou, Jun and Lin, Yankai and Wen, Ji-Rong and Li, Chongxuan},
  booktitle={Advances in Neural Information Processing Systems},
  year={2025},
  url={https://arxiv.org/abs/2502.09992}
}

@article{ye2025dream,
  title={Dream {7B}: Diffusion Large Language Models},
  author={Ye, Jiacheng and Xie, Zhihui and Zheng, Lin and Gao, Jiahui and Wu, Zirui and Jiang, Xin and Li, Zhenguo and Kong, Lingpeng},
  journal={arXiv preprint arXiv:2508.15487},
  year={2025},
  url={https://arxiv.org/abs/2508.15487}
}

@article{cobbe2021gsm8k,
  title={Training Verifiers to Solve Math Word Problems},
  author={Cobbe, Karl and Kosaraju, Vineet and Bavarian, Mohammad and Chen, Mark and Jun, Heewoo and Kaiser, Lukasz and Plappert, Matthias and Tworek, Jerry and Hilton, Jacob and Nakano, Reiichiro and Hesse, Christopher and Schulman, John},
  journal={arXiv preprint arXiv:2110.14168},
  year={2021},
  url={https://arxiv.org/abs/2110.14168}
}

@inproceedings{yu2023outcomesupervised,
  title={{OVM}, Outcome-supervised Value Models for Planning in Mathematical Reasoning},
  author={Yu, Fei and Gao, Anningzhe and Wang, Benyou},
  booktitle={Findings of the Association for Computational Linguistics: NAACL 2024},
  pages={858--875},
  year={2024},
  url={https://aclanthology.org/2024.findings-naacl.55/}
}

@inproceedings{yuan2024implicitprm,
  title={Free Process Rewards without Process Labels},
  author={Yuan, Lifan and Li, Wendi and Chen, Huayu and Cui, Ganqu and Ding, Ning and Zhang, Kaiyan and Zhou, Bowen and Liu, Zhiyuan and Peng, Hao},
  booktitle={International Conference on Machine Learning},
  year={2025},
  url={https://arxiv.org/abs/2412.01981}
}

@inproceedings{liao2025reward,
  title={Reward-Guided Speculative Decoding for Efficient {LLM} Reasoning},
  author={Liao, Baohao and Xu, Yuhui and Dong, Hanze and Li, Junnan and Monz, Christof and Savarese, Silvio and Sahoo, Doyen and Xiong, Caiming},
  booktitle={International Conference on Machine Learning},
  year={2025},
  url={https://arxiv.org/abs/2501.19324}
}

@inproceedings{kim2025fine,
  title={Fine-Tuning Masked Diffusion for Provable Self-Correction},
  author={Kim, Jaeyeon and Kim, Seunggeun and Lee, Taekyun and Pan, David Z. and Kim, Hyeji and Kakade, Sham M. and Chen, Sitan},
  booktitle={International Conference on Machine Learning},
  year={2026},
  url={https://arxiv.org/abs/2510.01384}
}

@inproceedings{bai2026prism,
  title={Prism: Efficient Test-Time Scaling via Hierarchical Search and Self-Verification for Discrete Diffusion Language Models},
  author={Bai, Jinbin and Li, Yixuan and Zhu, Yuchen and Xin, Yi and Shi, Qingyu and Feng, Aosong and Liu, Xiaohong and Tao, Molei and Xue, Jianru and Li, Xiangtai and Yang, Ming-Hsuan},
  booktitle={International Conference on Machine Learning},
  year={2026},
  url={https://arxiv.org/abs/2602.01842}
}

@article{luo2026self,
  title={Self-Rewarding Sequential {M}onte {C}arlo for Masked Diffusion Language Models},
  author={Luo, Ziwei and Jin, Ziqi and Wang, Lei and Bing, Lidong and Sch{\"o}n, Thomas B.},
  journal={arXiv preprint arXiv:2602.01849},
  year={2026},
  url={https://arxiv.org/abs/2602.01849}
}

@inproceedings{xie2026advancing,
  title={Advancing Reasoning in Diffusion Language Models with Denoising Process Rewards},
  author={Xie, Shaoan and Kong, Lingjing and Song, Xiangchen and Dong, Xinshuai and Chen, Guangyi and Xing, Eric P. and Zhang, Kun},
  booktitle={Proceedings of the 64th Annual Meeting of the Association for Computational Linguistics (Volume 1: Long Papers)},
  pages={42703--42720},
  year={2026},
  url={https://aclanthology.org/2026.acl-long.1978/}
}

@inproceedings{wang2025remasking,
  title={Remasking Discrete Diffusion Models with Inference-Time Scaling},
  author={Wang, Guanghan and Schiff, Yair and Sahoo, Subham Sekhar and Kuleshov, Volodymyr},
  booktitle={Advances in Neural Information Processing Systems},
  year={2025},
  url={https://arxiv.org/abs/2503.00307}
}

@inproceedings{hendrycks2021math,
title={Measuring Mathematical Problem Solving With the {MATH} Dataset},
author={Dan Hendrycks and Collin Burns and Saurav Kadavath and Akul Arora and Steven Basart and Eric Tang and Dawn Song and Jacob Steinhardt},
booktitle={Proceedings of the Neural Information Processing Systems Track on Datasets and Benchmarks},
year={2021},
url={https://arxiv.org/abs/2103.03874}
}

@inproceedings{lightman2023lets,
title={Let's Verify Step by Step},
author={Lightman, Hunter and Kosaraju, Vineet and Burda, Yura and Edwards, Harri and Baker, Bowen and Lee, Teddy and Leike, Jan and Schulman, John and Sutskever, Ilya and Cobbe, Karl},
booktitle={International Conference on Learning Representations},
year={2024},
url={https://arxiv.org/abs/2305.20050}
}

@article{chen2021codex,
title={Evaluating Large Language Models Trained on Code},
author={Mark Chen and Jerry Tworek and Heewoo Jun and Qiming Yuan and Henrique Ponde de Oliveira Pinto and Jared Kaplan and Harri Edwards and Yuri Burda and Nicholas Joseph and Greg Brockman and Alex Ray and Raul Puri and Gretchen Krueger and Michael Petrov and Heidy Khlaaf and Girish Sastry and Pamela Mishkin and Brooke Chan and Scott Gray and Nick Ryder and Mikhail Pavlov and Alethea Power and Lukasz Kaiser and Mohammad Bavarian and Clemens Winter and Philippe Tillet and Felipe Petroski Such and Dave Cummings and Matthias Plappert and Fotios Chantzis and Elizabeth Barnes and Ariel Herbert-Voss and William Hebgen Guss and Alex Nichol and Alex Paino and Nikolas Tezak and Jie Tang and Igor Babuschkin and Suchir Balaji and Shantanu Jain and William Saunders and Christopher Hesse and Andrew N. Carr and Jan Leike and Josh Achiam and Vedant Misra and Evan Morikawa and Alec Radford and Matthew Knight and Miles Brundage and Mira Murati and Katie Mayer and Peter Welinder and Bob McGrew and Dario Amodei and Sam McCandlish and Ilya Sutskever and Wojciech Zaremba},
journal={arXiv preprint arXiv:2107.03374},
year={2021},
eprint={2107.03374},
archivePrefix={arXiv},
primaryClass={cs.LG},
url={https://arxiv.org/abs/2107.03374}
}

@article{austin2021program,
title={Program Synthesis with Large Language Models},
author={Austin, Jacob and Odena, Augustus and Nye, Maxwell and Bosma, Maarten and Michalewski, Henryk and Dohan, David and Jiang, Ellen and Cai, Carrie and Terry, Michael and Le, Quoc and others},
journal={arXiv preprint arXiv:2108.07732},
year={2021},
url={https://arxiv.org/abs/2108.07732}
}

@inproceedings{zhao2021calibrating,
  title={Calibrating Predictions to Decisions: A Novel Approach to Multi-Class Calibration},
  author={Zhao, Shengjia and Kim, Michael P. and Sahoo, Roshni and Ma, Tengyu and Ermon, Stefano},
  booktitle={Advances in Neural Information Processing Systems},
  volume={34},
  year={2021},
  url={https://arxiv.org/abs/2107.05719}
}

@inproceedings{sahoo2024simple,
  title={Simple and Effective Masked Diffusion Language Models},
  author={Sahoo, Subham Sekhar and Arriola, Marianne and Schiff, Yair and Gokaslan, Aaron and Marroquin, Edgar and Chiu, Justin T and Rush, Alexander and Kuleshov, Volodymyr},
  booktitle={Advances in Neural Information Processing Systems},
  year={2024},
  url={https://arxiv.org/abs/2406.07524}
}

@article{khanna2025mercury,
  title={Mercury: Ultra-Fast Language Models Based on Diffusion},
  author={{Inception Labs} and Khanna, Samar and Kharbanda, Siddhant and Li, Shufan and Varma, Harshit and Wang, Eric and Birnbaum, Sawyer and Luo, Ziyang and Miraoui, Yanis and Palrecha, Akash and Ermon, Stefano and Grover, Aditya and Kuleshov, Volodymyr},
  journal={arXiv preprint arXiv:2506.17298},
  year={2025},
  url={https://arxiv.org/abs/2506.17298}
}

@inproceedings{wu2026fast,
  title={{Fast-dLLM}: Training-free Acceleration of Diffusion {LLM} by Enabling {KV} Cache and Parallel Decoding},
  author={Wu, Chengyue and Zhang, Hao and Xue, Shuchen and Liu, Zhijian and Diao, Shizhe and Zhu, Ligeng and Luo, Ping and Han, Song and Xie, Enze},
  booktitle={International Conference on Learning Representations},
  year={2026},
  url={https://arxiv.org/abs/2505.22618}
}

@inproceedings{ye2025beyond,
  title={Beyond Autoregression: Discrete Diffusion for Complex Reasoning and Planning},
  author={Ye, Jiacheng and Gao, Jiahui and Gong, Shansan and Zheng, Lin and Jiang, Xin and Li, Zhenguo and Kong, Lingpeng},
  booktitle={International Conference on Learning Representations},
  year={2025},
  url={https://arxiv.org/abs/2410.14157}
}

@article{peng2025path,
  title={Path Planning for Masked Diffusion Model Sampling},
  author={Peng, Fred Zhangzhi and Bezemek, Zachary and Patel, Sawan and Rector-Brooks, Jarrid and Yao, Sherwood and Bose, Avishek Joey and Tong, Alexander and Chatterjee, Pranam},
  journal={arXiv preprint arXiv:2502.03540},
  year={2025},
  url={https://arxiv.org/abs/2502.03540}
}

@inproceedings{hong2026wide,
  title={Wide-In, Narrow-Out: Revokable Decoding for Efficient and Effective {DLLMs}},
  author={Hong, Feng and Yu, Geng and Ye, Yushi and Huang, Haicheng and Zheng, Huangjie and Zhang, Ya and Wang, Yanfeng and Yao, Jiangchao},
  booktitle={International Conference on Learning Representations},
  year={2026},
  url={https://arxiv.org/abs/2507.18578}
}

@article{zhao2026lockr,
  title={{LOCKR}: A Hidden-State Trajectory-Guided Planner for Detecting and Repairing Stable-but-Wrong Lock-In in Diffusion Language Models},
  author={Zhao, Guoshenghui and Yu, Tan and Zhao, Weijie},
  journal={arXiv preprint arXiv:2609.27220},
  year={2026},
  url={https://arxiv.org/abs/2609.27220}
}

@inproceedings{huang2026remedi,
  title={Don't Settle Too Early: Self-Reflective Remasking for Diffusion Language Models},
  author={Huang, Zemin and Wang, Yuhang and Chen, Zhiyang and Qi, Guo-Jun},
  booktitle={International Conference on Learning Representations},
  year={2026},
  url={https://arxiv.org/abs/2509.23653}
}

@inproceedings{jazbec2026learning,
  title={Learning Unmasking Policies for Diffusion Language Models},
  author={Jazbec, Metod and Olausson, Theo X. and B{\'e}thune, Louis and Ablin, Pierre and Kirchhof, Michael and Monteiro, Jo{\~a}o and Turrisi, Victor and Ramapuram, Jason and Cuturi, Marco},
  booktitle={International Conference on Machine Learning},
  year={2026},
  url={https://arxiv.org/abs/2512.09106}
}

@inproceedings{vonrutte2025gidd,
  title={Generalized Interpolating Discrete Diffusion},
  author={von R{\"u}tte, Dimitri and Fluri, Janis and Ding, Yuhui and Orvieto, Antonio and Sch{\"o}lkopf, Bernhard and Hofmann, Thomas},
  booktitle={International Conference on Machine Learning},
  year={2025},
  url={https://arxiv.org/abs/2503.04482}
}

@inproceedings{zhao2025d1,
  title={d1: Scaling Reasoning in Diffusion Large Language Models via Reinforcement Learning},
  author={Zhao, Siyan and Gupta, Devaansh and Zheng, Qinqing and Grover, Aditya},
  booktitle={Advances in Neural Information Processing Systems},
  year={2025},
  url={https://arxiv.org/abs/2504.12216}
}

@inproceedings{orgad2025llms,
  title={{LLMs} Know More Than They Show: On the Intrinsic Representation of {LLM} Hallucinations},
  author={Orgad, Hadas and Toker, Michael and Gekhman, Zorik and Reichart, Roi and Szpektor, Idan and Kotek, Hadas and Belinkov, Yonatan},
  booktitle={International Conference on Learning Representations},
  year={2025},
  url={https://arxiv.org/abs/2410.02707}
}

@inproceedings{li2026diffusion,
  title={Diffusion Language Model Knows the Answer Before It Decodes},
  author={Li, Pengxiang and Zhou, Yefan and Muhtar, Dilxat and Yin, Lu and Yan, Shilin and Shen, Li and Liang, Yi and Vosoughi, Soroush and Liu, Shiwei},
  booktitle={International Conference on Learning Representations},
  year={2026},
  url={https://arxiv.org/abs/2508.19982}
}

@inproceedings{wang2026time,
  title={Time Is a Feature: Exploiting Temporal Dynamics in Diffusion Language Models},
  author={Wang, Wen and Fang, Bozhen and Jing, Chenchen and Shen, Yongliang and Shen, Yangyi and Wang, Qiuyu and Ouyang, Hao and Chen, Hao and Shen, Chunhua},
  booktitle={International Conference on Learning Representations},
  year={2026},
  url={https://arxiv.org/abs/2508.09138}
}

@inproceedings{chiu2026probing,
  title={Probing Functional Correctness in Diffusion Language Models},
  author={Chiu, Guan-Ming and Liu, Jeng-Yue},
  booktitle={Proceedings of the 64th Annual Meeting of the Association for Computational Linguistics (Volume 4: Student Research Workshop)},
  pages={163--172},
  year={2026},
  url={https://aclanthology.org/2026.acl-srw.15/}
}

@inproceedings{zhang2025corrective,
  title={Corrective Diffusion Language Models},
  author={Zhang, Shuibai and Peng, Fred Zhangzhi and Zhang, Yiheng and Pan, Jin and Chrysos, Grigorios G.},
  booktitle={Advances in Neural Information Processing Systems},
  year={2026},
  url={https://arxiv.org/abs/2512.15596}
}

@inproceedings{hu2022lora,
  title={{LoRA}: Low-Rank Adaptation of Large Language Models},
  author={Hu, Edward J. and Shen, Yelong and Wallis, Phillip and Allen-Zhu, Zeyuan and Li, Yuanzhi and Wang, Shean and Wang, Lu and Chen, Weizhu},
  booktitle={International Conference on Learning Representations},
  year={2022},
  url={https://arxiv.org/abs/2106.09685}
}

@incollection{platt1999probabilistic,
  title={Probabilistic Outputs for Support Vector Machines and Comparisons to Regularized Likelihood Methods},
  author={Platt, John C.},
  booktitle={Advances in Large Margin Classifiers},
  editor={Smola, Alexander J. and Bartlett, Peter L. and Sch{\"o}lkopf, Bernhard and Schuurmans, Dale},
  publisher={MIT Press},
  pages={61--74},
  year={1999}
}

@misc{pan2025tinyzero,
  title={{TinyZero}},
  author={Pan, Jiayi and Zhang, Junjie and Wang, Xingyao and Yuan, Lifan and Peng, Hao and Suhr, Alane},
  howpublished={\url{https://github.com/Jiayi-Pan/TinyZero}},
  year={2025}
}

@inproceedings{mounier2025r3,
  title={Review, Remask, Refine ({R3}): Process-Guided Block Diffusion for Text Generation},
  author={Mounier, Nikita and Idehpour, Parsa},
  booktitle={ICML 2025 Workshop on Methods and Opportunities at Small Scale},
  year={2025},
  url={https://arxiv.org/abs/2507.08018}
}

@inproceedings{dang2026inference,
  title={Inference-Time Scaling of Diffusion Language Models via Trajectory Refinement},
  author={Dang, Meihua and Han, Jiaqi and Xu, Minkai and Xu, Kai and Srivastava, Akash and Ermon, Stefano},
  booktitle={Conference on Language Modeling},
  year={2026},
  url={https://arxiv.org/abs/2507.08390}
}

@inproceedings{misaki2026unmaskfork,
  title={{UnMaskFork}: Test-Time Scaling for Masked Diffusion via Deterministic Action Branching},
  author={Misaki, Kou and Akiba, Takuya},
  booktitle={International Conference on Machine Learning},
  year={2026},
  url={https://arxiv.org/abs/2602.04344}
}

@inproceedings{miles2026stitching,
  title={Test-Time Scaling with Diffusion Language Models via Reward-Guided Stitching},
  author={Miles, Roy and Toker, Aysim and Oncescu, Andreea-Maria and Deng, Jiankang and Elezi, Ismail},
  booktitle={Advances in Neural Information Processing Systems},
  year={2026},
  url={https://arxiv.org/abs/2602.22871}
}

@inproceedings{zhan2026why,
  title={Why Deterministic {PRM} Guidance Underperforms in Discrete Diffusion Reasoning},
  author={Zhan, Yan and Liu, Shaobo and Gao, Zhijun},
  booktitle={Advances in Neural Information Processing Systems},
  year={2026},
  url={https://arxiv.org/abs/2609.35472}
}

@article{jeon2026vgb,
  title={{VGB} for Masked Diffusion Model: Efficient Test-time Scaling for Reward Satisfaction and Sample Editing},
  author={Jeon, Kijung and Vuong, Thuy-Duong and Tao, Molei},
  journal={arXiv preprint arXiv:2606.28301},
  year={2026},
  url={https://arxiv.org/abs/2606.28301}
}
\newpage{}

\appendix

\section{Diagnosis Details}
\label{sec:app-diagnosis}
All Sudoku results in the main text use the 190 uniquely solvable puzzles from a fixed 256-puzzle sample of the d1 $4{\times}4$ test set \citep{zhao2025d1}. Overall solve rates, probe datasets, localization subsets, and repair-eligible failures have different denominators. The final-state probe analyses contain 1,416 eligible cells for remasking and 1,279 for commit-once, distinct from the 190-puzzle solve-rate denominator. Single-layer probes use standardization, PCA, and class-balanced logistic regression with five-fold cross-validation grouped by puzzle. Bootstrap intervals resample puzzles rather than cells, because repeated cell observations from one puzzle are not independent. The reported bootstrap intervals condition on fitted out-of-fold probe predictions, and they do not rerun probe training in each resample. For the remasking probe they are $[0.793, 0.881]$ at the final state and $[0.836, 0.892]$ at first commit, from 500 resamples. Probes read answer-position hidden states at layers 8, 16, 24 and 32 at every step, excluding given clues and echoed blank digits, and reported AUROCs pool out-of-fold cell scores. The within-puzzle detector of Finding 1 is trained with puzzle-grouped folds on decoding states from the first quarter of decoding onward and is distinct from the single-layer probes. On the 73 remasking final states that contain both correct and incorrect eligible cells, it reaches a mean within-puzzle AUROC of 0.817, against 0.489 for confidence, and its median number of correct cells ranked before the last erroneous cell is one, against three for confidence. The gold-guided repair of Finding 2 uses the 77 of the 84 unsolved remasking outputs that have eligible erroneous answer cells. Rollback depth $d\in\{0,8,16,32,64\}$ additionally reopens the non-answer tokens first committed at or after step $\max(0,\tau-d)$, where $\tau$ is the earliest first-commit step among the final wrong cells. The one-cell re-prediction summary of Finding 2 pools 1,292 wrong-commit observations from three balanced rounds at offset zero, not 1,292 independent puzzles, and observation counts vary across offsets, so the comparison across offsets is descriptive rather than a paired estimate of how each error evolves.

\section{Verifier Training and Evaluation}
\label{sec:app-verifier}
\paragraph{Verification and generation.}
The primary verifier uses four samples for each of 2,000 GSM8K training questions, decoded with the canvas, block length and step schedule of the main experiments. Every verifier uses a rank-32 adapter with scaling $\alpha=64$ on the attention and feed-forward projections, and its head maps the hidden dimension through 1,024 SiLU units to a single logit. Training runs for two epochs with gradients accumulated over four questions. It pools the full response canvas, including unresolved masks. An already-terminated sample contributes its final boundary when a later training index is drawn. For the SFT generator, the SFT adapter is merged into the backbone before the verifier adapter is added, so that disabling the verifier adapter leaves exactly the SFT generator. Pair accuracy averages correct--incorrect ranking accuracy within each problem, then across problems with both outcomes, and ties receive half credit. After block $b$, a candidate that has already terminated contributes its final state, and 488 of the 1,319 GSM8K test problems have both outcomes among their four candidates. For each guided run, the paired greedy baseline uses the same process and decoding configuration. Cost counts forward passes weighted by batch rows.

\paragraph{Seed sensitivity.}
The primary GSM8K run fixes 154 failures and breaks 20 correct answers. Changing only the verifier-training seed gives 77.5\% success, and changing only the re-decoding seed gives 77.6\%, against 78.8\% in the primary run and 68.7\% for paired greedy decoding. These are separate sensitivity checks rather than independent end-to-end replications.

\paragraph{Evaluation subsets.}
MATH-500 is the 500-problem subset of MATH used in process-verification research \citep{hendrycks2021math,lightman2023lets}, and the implementation loads its public dataset release. For Countdown, the sampler filters three-number tasks from the training partition of the public Countdown-Tasks-3to4 collection \citep{pan2025tinyzero} and selects 500 for evaluation.

\paragraph{Code evaluation.}
The code results use all 500 MBPP test tasks and 164 HumanEval tasks. MBPP prompts expose the same tests used by its grader. HumanEval runs its test harness on the prompt and extracted program. Programs have a 10-second timeout and a 2-GiB address-space limit, and extraction or execution failures count as incorrect.

\paragraph{Regeneration locations.}
The repair block of Table~\ref{tab:regen-blocks} is the first block at which the guided decode's recorded score departs from the paired greedy decode's, which identifies the first accepted alternative exactly.

\section{Baselines}
\label{sec:app-baselines}
\paragraph{Baseline implementations.}
SR-SMC is a local blockwise implementation with four particles at temperature 1.0, resampled after a block when the effective sample size falls below two, with the answer taken from the particle of highest total commitment log-confidence.

The Prism port keeps the paper's hierarchical search, self-verified pruning and remasking.
\begin{itemize}
\item \textbf{Search.} Sixteen initial trajectories are pruned to $K{=}4$ within the first block's 10--60\% step window, every three steps (decay 1.8). Two survivors branch with two low-confidence tokens remasked. The generator's own Yes/No judgment scores the trajectories.
\item \textbf{Answer.} As in the main results of \citet{bai2026prism} (their Algorithm~1), the answer is a majority vote over the $K$ final trajectories, for every benchmark including code. The released code instead returns the final trajectory with the highest judgment score, and Table~\ref{tab:prism-variants} reports that variant.
\item \textbf{Cost.} Rows count every judgment read used for pruning. The final judgment reads that the vote does not need are excluded.
\end{itemize}

\begin{table}[t]
\caption{\textbf{Prism answer selection.} Vote is the majority over the final trajectories, as in \citet{bai2026prism}, and is the number in Table~\ref{tab:main-results}. Top is the released code's choice, the trajectory with the highest self-verification score. Cost is that of the vote, including every pruning judgment.}
\label{tab:prism-variants}
\centering
\small
\setlength{\tabcolsep}{4pt}
\renewcommand{\arraystretch}{1.08}
\begin{tabular*}{\textwidth}{@{\extracolsep{\fill}}lrrrrrrrrrrrr@{}}
\toprule
& \multicolumn{4}{c}{LLaDA-8B-Base + SFT} & \multicolumn{4}{c}{LLaDA-8B-Instruct} & \multicolumn{4}{c}{Dream-v0-Instruct-7B} \\
\cmidrule(lr){2-5}\cmidrule(lr){6-9}\cmidrule(l){10-13}
Benchmark & Greedy & Vote & Top & Cost & Greedy & Vote & Top & Cost & Greedy & Vote & Top & Cost \\
\midrule
GSM8K & 68.7 & 71.9 & 67.6 & $\times5.9$ & 79.2 & 83.5 & 82.8 & $\times5.0$ & 77.3 & 85.1 & 83.1 & $\times5.5$ \\
MATH-500 & 24.6 & 32.4 & 29.0 & $\times6.0$ & 35.0 & 43.4 & 38.0 & $\times4.8$ & 40.8 & 48.2 & 49.6 & $\times5.4$ \\
Countdown & 35.6 & 39.8 & 40.8 & $\times7.0$ & 15.6 & 21.2 & 27.4 & $\times4.9$ & 18.0 & 29.2 & 46.2 & $\times5.0$ \\
HumanEval & 23.2 & 20.1 & 25.0 & $\times5.2$ & 34.1 & 28.0 & 36.0 & $\times5.1$ & 45.1 & 42.7 & 53.7 & $\times5.9$ \\
MBPP & 24.2 & 22.4 & 28.4 & $\times7.4$ & 33.2 & 31.6 & 36.6 & $\times5.6$ & 43.8 & 46.4 & 53.2 & $\times7.5$ \\
\bottomrule
\end{tabular*}
\end{table}

\paragraph{Signal-control details.}
Commitment confidence is the geometric mean of token probabilities recorded when the visible tokens were placed. Prompted self-verification uses the same generator, a judgment prompt and a masked readout position, scoring Yes against No token variants. Each signal's threshold maximizes accuracy on its 252-problem development evaluation, namely 0.85 for confidence, 0.98 for Yes/No, 0.7 for V0, and 0.5 for DiVR. All test runs cover 1,319 GSM8K problems.

\paragraph{Selection baselines.}
The completed mean-log-probability selector re-reads samples with their response tokens visible, so it is not a normalized joint sequence likelihood. Countdown majority voting groups boxed expressions after whitespace removal rather than by their value, which would merge distinct expressions.

\clearpage
\section{Additional Generators and Verifiers}
\label{sec:app-generators}
\paragraph{GSM8K SFT of the instruction-tuned generators.} \begin{table}[t]
\caption{\textbf{GSM8K SFT of the instruction-tuned generators.} Same SFT data and recipe as the LLaDA-8B-Base SFT generator, each with a verifier trained on its own decodes. All verifiers here are GSM8K-trained, code columns included.}
\label{tab:inst-sft}
\centering
\small
\setlength{\tabcolsep}{2.5pt}
\renewcommand{\arraystretch}{1.08}
\begin{tabular*}{\textwidth}{@{\extracolsep{\fill}}l*{5}{r@{\hspace{4pt}}l}@{}}
\toprule
\textbf{Generator / method} & \multicolumn{2}{c}{GSM8K} & \multicolumn{2}{c}{MATH-500} & \multicolumn{2}{c}{Countdown} & \multicolumn{2}{c}{HumanEval} & \multicolumn{2}{c}{MBPP} \\
\cmidrule(lr){2-3}\cmidrule(lr){4-5}\cmidrule(lr){6-7}\cmidrule(lr){8-9}\cmidrule(l){10-11}
& \multicolumn{1}{c}{Acc.} & Cost & \multicolumn{1}{c}{Acc.} & Cost & \multicolumn{1}{c}{Acc.} & Cost & \multicolumn{1}{c}{Acc.} & Cost & \multicolumn{1}{c}{Acc.} & Cost \\
\midrule
\multicolumn{11}{@{}l}{\textbf{LLaDA-8B-Instruct}} \\
\quad Released, greedy & 79.2 & $\times1.0$ & 35.0 & $\times1.0$ & 15.6 & $\times1.0$ & 34.1 & $\times1.0$ & 33.2 & $\times1.0$ \\
\quad Released + DiVR & 83.0 & $\times1.6$ & 38.2 & $\times1.8$ & 23.0 & $\times2.6$ & 34.8 & $\times2.6$ & 38.0 & $\times2.6$ \\
\quad + GSM8K SFT, greedy & 74.7 & $\times1.0$ & 30.8 & $\times1.0$ & 11.4 & $\times1.0$ & 26.2 & $\times1.0$ & 28.8 & $\times1.0$ \\
\quad + GSM8K SFT + DiVR & 81.6 & $\times1.9$ & 37.0 & $\times1.8$ & 42.2 & $\times2.7$ & 26.8 & $\times2.3$ & 31.6 & $\times2.2$ \\
\midrule
\multicolumn{11}{@{}l}{\textbf{Dream-v0-Instruct-7B}} \\
\quad Released, greedy & 77.3 & $\times1.0$ & 40.8 & $\times1.0$ & 18.0 & $\times1.0$ & 45.1 & $\times1.0$ & 43.8 & $\times1.0$ \\
\quad Released + DiVR & 84.4 & $\times1.7$ & 44.4 & $\times1.9$ & 37.2 & $\times2.4$ & 49.4 & $\times2.6$ & 50.0 & $\times3.4$ \\
\quad + GSM8K SFT, greedy & 67.1 & $\times1.0$ & 33.2 & $\times1.0$ & 20.8 & $\times1.0$ & 41.5 & $\times1.0$ & 44.8 & $\times1.0$ \\
\quad + GSM8K SFT + DiVR & 79.6 & $\times1.8$ & 40.4 & $\times1.7$ & 31.0 & $\times2.7$ & 43.3 & $\times1.7$ & 44.2 & $\times2.1$ \\
\bottomrule
\end{tabular*}
\end{table}Table~\ref{tab:inst-sft} applies the LLaDA-8B-Base SFT recipe unchanged to LLaDA-8B-Instruct and Dream-v0-Instruct-7B (for Dream, the loss reads each position's prediction from the previous output, matching its decoding convention). SFT lowers both generators. GSM8K greedy accuracy falls from 79.2\% to 74.7\% for Instruct and from 77.3\% to 67.1\% for Dream, and the responses shorten, consistent with imitation of the shorter reference solutions. Guidance with each SFT generator's own verifier brings it to within 1.8 points of the released greedy accuracy or above it on nine of the ten generator--benchmark pairs (the exception is Instruct on HumanEval), but the released generator with its verifier stays ahead on all five benchmarks for Dream and on four of five for Instruct, where only Countdown favors the SFT generator (42.2\% against 23.0\%). Table~\ref{tab:main-results} therefore uses the released checkpoints for these two generators.

\paragraph{diffu-GRPO reproduction.}
\begin{table}[t]
\caption{\textbf{diffu-GRPO generators under every method.} LLaDA-8B-Instruct and the untuned LLaDA-8B-Base after 3{,}800 steps of d1's diffu-GRPO recipe on GSM8K, each with DiVR verifiers trained on its own decodes. The greedy rows of the untrained generators repeat Table~\ref{tab:main-results}. $^\dagger$Local reimplementations.}
\label{tab:rl}
\centering
\small
\setlength{\tabcolsep}{2.5pt}
\renewcommand{\arraystretch}{1.08}
\begin{tabular*}{\textwidth}{@{\extracolsep{\fill}}l*{5}{r@{\hspace{4pt}}l}r@{}}
\toprule
\textbf{Model / method} & \multicolumn{6}{c}{\textbf{Math}} & \multicolumn{4}{c}{\textbf{Code}} & \\
\cmidrule(lr){2-7}\cmidrule(l){8-11}
& \multicolumn{2}{c}{GSM8K} & \multicolumn{2}{c}{MATH-500} & \multicolumn{2}{c}{Countdown} & \multicolumn{2}{c}{HumanEval} & \multicolumn{2}{c}{MBPP} & \multicolumn{1}{c}{Avg.} \\
\cmidrule(lr){2-3}\cmidrule(lr){4-5}\cmidrule(lr){6-7}\cmidrule(lr){8-9}\cmidrule(lr){10-11}\cmidrule(l){12-12}
& \multicolumn{1}{c}{Acc.} & Cost & \multicolumn{1}{c}{Acc.} & Cost & \multicolumn{1}{c}{Acc.} & Cost & \multicolumn{1}{c}{Acc.} & Cost & \multicolumn{1}{c}{Acc.} & Cost & \multicolumn{1}{c}{Acc.} \\
\midrule
\multicolumn{12}{@{}l}{\textbf{LLaDA-8B-Instruct}} \\
\quad Greedy (Table~\ref{tab:main-results}) & 79.2 & $\times1.0$ & 35.0 & $\times1.0$ & 15.6 & $\times1.0$ & 34.1 & $\times1.0$ & 33.2 & $\times1.0$ & 39.4 \\
\quad + diffu-GRPO & 77.4 & $\times1.0$ & 39.4 & $\times1.0$ & 17.0 & $\times1.0$ & 31.1 & $\times1.0$ & 35.4 & $\times1.0$ & 40.1 \\
\quad + diffu-GRPO + SR-SMC (4)$^\dagger$ & 81.2 & $\times4.1$ & 37.2 & $\times4.0$ & 23.6 & $\times3.9$ & 32.3 & $\times4.0$ & 37.6 & $\times4.1$ & 42.4 \\
\quad + diffu-GRPO + Prism$^\dagger$ & 84.4 & $\times5.2$ & 41.4 & $\times4.9$ & 22.6 & $\times5.1$ & 28.0 & $\times5.1$ & 33.4 & $\times5.6$ & 42.0 \\
\quad + diffu-GRPO + DiVR-guided & 83.4 & $\times1.6$ & 39.2 & $\times1.8$ & 28.2 & $\times2.7$ & 34.8 & $\times2.7$ & 41.4 & $\times2.3$ & 45.4 \\
\quad + diffu-GRPO + DiVR best-of-4 & 87.4 & $\times4.0$ & 44.8 & $\times3.9$ & 42.0 & $\times4.0$ & 36.0 & $\times4.0$ & 42.0 & $\times4.0$ & 50.4 \\
\midrule
\multicolumn{12}{@{}l}{\textbf{LLaDA-8B-Base}} \\
\quad Greedy (Table~\ref{tab:main-results}) & 61.8 & $\times1.0$ & 24.2 & $\times1.0$ & 10.6 & $\times1.0$ & 19.5 & $\times1.0$ & 30.2 & $\times1.0$ & 29.3 \\
\quad + diffu-GRPO & 61.7 & $\times1.0$ & 21.6 & $\times1.0$ & 15.6 & $\times1.0$ & 23.8 & $\times1.0$ & 26.0 & $\times1.0$ & 29.7 \\
\quad + diffu-GRPO + DiVR-guided & 76.2 & $\times1.8$ & 26.4 & $\times1.8$ & 31.6 & $\times2.6$ & 26.8 & $\times2.1$ & 35.6 & $\times2.0$ & 39.3 \\
\quad + diffu-GRPO + DiVR best-of-4 & 82.7 & $\times4.0$ & 31.6 & $\times4.0$ & 38.0 & $\times4.0$ & 31.1 & $\times4.0$ & 35.8 & $\times4.4$ & 43.8 \\
\bottomrule
\end{tabular*}
\end{table}
The Instruct RL generator of Tables~\ref{tab:main-results} and~\ref{tab:rl} is LLaDA-8B-Instruct trained with d1's released diffu-GRPO code and GSM8K reward functions \citep{zhao2025d1}, with LoRA rank 128 ($\alpha{=}64$, dropout 0.05) on the attention and feed-forward projections over a 4-bit base, six generations per prompt at temperature 0.9, 256-token completions in blocks of 32 with 128 diffusion steps, twelve inner updates per batch, learning rate $3\times10^{-6}$, KL coefficient 0.04, clipping 0.5, seed 42, and 3{,}800 optimizer steps with one prompt per GPU and no gradient accumulation on four RTX A6000 GPUs (12.0 hours). d1 trains its GSM8K model for 7{,}700 steps on eight A100 GPUs with two accumulation steps, so our training uses about one eighth as many generations, and its correctness reward was still rising slowly at the end.

\emph{Greedy accuracy.} With the single-problem greedy decoder used for the released checkpoints, the trained generator scores 78.8\%, 36.4\% and 18.8\% on GSM8K, MATH-500 and Countdown against 78.5\%, 35.6\% and 17.6\% for Instruct ($p=0.84$, $0.76$ and $0.66$). In the paired decoder of the guided runs it scores 77.4\%, 39.4\% and 17.0\% against 79.2\%, 35.0\% and 15.6\% (Table~\ref{tab:rl}). The two greedy decoders differ by up to 3.0 points on MATH-500 for the same adapter, which is why every comparison in this paper is paired within one decoder.

\emph{Comparison with d1.} In d1's own evaluation (512 tokens, 256 steps, d1's prompt and parser) it scores 80.1\% against 79.0\% for Instruct ($p=0.34$). A second run of the same recipe and seed scored 81.1\% in that evaluation ($p=0.058$), and the two runs differ on 195 problems with no net direction ($p=0.39$). d1's authors released no LLaDA checkpoint. The public third-party adapter \texttt{dLLM-data-selection/\allowbreak LlaDA-8B-\allowbreak Instruct-\allowbreak grpo-\allowbreak gsm8k-\allowbreak pass-\allowbreak rate-\allowbreak step-1000} (diffu-GRPO, 1{,}000 steps) scores 78.8\%, 36.0\% and 15.2\% in our evaluation.

\emph{Reward.} Re-grading the samples generated during training, d1's exact-string correctness rose from 24\% to 71\% of samples over training while our numeric grader rose from 76\% to 79\%, so much of the reward increase reflects the answer format that d1's reward and parser require.

\emph{Verifiers.} Its verifiers repeat the Instruct recipes on its own decodes (82.3\% of the GSM8K samples correct, held-out pair accuracy 0.833 and 0.885 with the second epoch kept, final-state pair accuracy 0.881 on the test pools, and for MBPP 35.8\% passing with 0.822 and 0.844).

\emph{Results.} Table~\ref{tab:rl} runs every method of Table~\ref{tab:main-results} on this generator. Against Instruct's greedy decoding, its greedy decoding differs by $-1.8$, $+4.4$, $+1.4$, $-3.0$ and $+2.2$ points on the five benchmarks (McNemar $p=0.15$, $0.025$, $0.57$, $0.23$ and $0.14$). SR-SMC and Prism move it by about as much as they move Instruct. With the same method on both generators, the only significant differences are SR-SMC on MATH-500 ($-4.0$, $p=0.0495$) and MBPP ($+3.6$, $p=0.047$) and best-of-4 on Countdown ($+6.6$, $p=0.019$).

\paragraph{diffu-GRPO from the untuned Base model.}
The Base RL generator applies the same recipe, seed and step count to the untuned LLaDA-8B-Base with two changes that its cold start requires.

\emph{Prompt prefix.} The prompt ends with the \texttt{<reasoning>} prefix that every Base decode in this paper uses (Appendix~\ref{sec:prompts}), passed to d1's trainer as an opening assistant turn so that the rendered prompt is identical to ours. Without the prefix the untuned model does not produce d1's answer tags and receives no correctness reward.

\emph{Correctness reward.} The correctness reward is our numeric grader (the first \texttt{<answer>} block, else a \texttt{\#\#\#\#} line, else the last boxed expression, else the last number, compared numerically) in place of d1's exact string match, with d1's four format rewards unchanged. The untuned model writes sentences inside the answer tags, so the exact match rewards 4\% of its samples where our grader scores 53\% right, and groups of six generations then carry almost no reward variance for the policy gradient to use.

\emph{Training.} Training took 52 GPU-hours on the same four GPUs, its correctness reward rose from 1.03 to 1.19 of 2, and its completions fill the 256-token canvas because the Base model writes no end token, as in every untuned-Base decode of the paper. The Instruct run keeps d1's reward.

\emph{Results.} The Base RL generator's greedy accuracy stays at the untuned level, 29.7\% against 29.3\% averaged over the five benchmarks and 61.7\% against 61.8\% on GSM8K (199 problems gained and 200 lost), with $+5.0$ on Countdown ($p=0.022$), $+4.3$ on HumanEval ($p=0.09$), $-2.6$ on MATH-500 and $-4.2$ on MBPP ($p=0.022$), and 5.6 points below one GPU-hour of GSM8K SFT on average (Table~\ref{tab:rl}).

\emph{Verifiers.} Its verifiers repeat the recipes of this appendix on its own decodes (65.1\% of the GSM8K samples correct, held-out pair accuracy 0.938 and 0.949, final-state pair accuracy 0.927 on the test pools against 0.919 for the SFT generator's verifier, and for MBPP 24.4\% passing with 0.776 and 0.908).

\emph{Guidance.} With guidance it gains 14.5 points on GSM8K, 16.0 on Countdown, 4.8 on MATH-500 ($p=0.015$), 9.6 on MBPP and 3.0 on HumanEval ($p=0.06$), its GSM8K guidance is 4.5 points above the SFT verifier's guidance of the untuned generator ($p=0.0005$), and best-of-4 reaches 82.7\% on GSM8K, the level of the SFT generator with its verifier (82.2\%). The lenient reward function is included with our code.

\paragraph{Code verifiers.} \begin{table}[t]
\caption{\textbf{Code benchmarks with the GSM8K-trained and the MBPP-trained verifier.} The two verifiers of a generator differ only in their training decodes, and Table~\ref{tab:main-results} reports the MBPP verifier in its code columns. Pair is the within-problem accuracy of ranking a passing finished sample above a failing one. $^\ast$Verified by the verifiers of the SFT generator.}
\label{tab:code-transfer}
\centering
\small
\setlength{\tabcolsep}{4pt}
\renewcommand{\arraystretch}{1.08}
\begin{tabular*}{\textwidth}{@{\extracolsep{\fill}}lrrrrrrrr@{}}
\toprule
& \multicolumn{4}{c}{HumanEval} & \multicolumn{4}{c}{MBPP} \\
\cmidrule(lr){2-5}\cmidrule(l){6-9}
\textbf{Generator / verifier} & Guided & Cost & Best-of-4 & Pair & Guided & Cost & Best-of-4 & Pair \\
\midrule
\multicolumn{9}{@{}l}{\textbf{LLaDA-8B-Base}, untuned$^\ast$ (greedy 19.5 / 30.2)} \\
\quad GSM8K verifier & 22.0 & $\times2.0$ & 27.4 & 0.778 & 30.2 & $\times2.0$ & 30.2 & 0.659 \\
\quad MBPP verifier & 23.8 & $\times2.7$ & 20.7 & 0.616 & 37.6 & $\times2.8$ & 33.4 & 0.757 \\
\multicolumn{9}{@{}l}{\textbf{LLaDA-8B-Base + SFT} (greedy 23.2 / 24.2)} \\
\quad GSM8K verifier & 24.4 & $\times2.1$ & 24.4 & 0.813 & 24.2 & $\times2.0$ & 26.8 & 0.759 \\
\quad MBPP verifier & 23.8 & $\times2.9$ & 26.2 & 0.838 & 33.6 & $\times3.3$ & 32.6 & 0.936 \\
\multicolumn{9}{@{}l}{\textbf{LLaDA-8B-Instruct} (greedy 34.1 / 33.2)} \\
\quad GSM8K verifier & 34.8 & $\times2.6$ & 36.6 & 0.774 & 38.0 & $\times2.6$ & 40.0 & 0.823 \\
\quad MBPP verifier & 39.0 & $\times2.6$ & 40.2 & 0.944 & 40.8 & $\times2.5$ & 40.2 & 0.867 \\
\multicolumn{9}{@{}l}{\textbf{Dream-v0-Instruct-7B} (greedy 45.1 / 43.8)} \\
\quad GSM8K verifier & 49.4 & $\times2.6$ & 53.7 & 0.859 & 50.0 & $\times3.4$ & 51.2 & 0.839 \\
\quad MBPP verifier & 53.0 & $\times2.0$ & 57.3 & 0.942 & 55.2 & $\times2.9$ & 55.0 & 0.943 \\
\bottomrule
\end{tabular*}
\end{table}Each generator's code verifier repeats the recipe of its GSM8K verifier on four samples at temperature 0.7 for each of the 464 MBPP training and validation problems. Sixty-four problems are held out, and the epoch with the higher held-out pair accuracy is kept (0.918, 0.909 and 0.953 for the SFT generator, Instruct and Dream). On the four-sample test pools of the three generators with their own verifiers, the GSM8K verifiers rank passing above failing programs with pair accuracy 0.759--0.859 and the MBPP verifiers with 0.838--0.944 (Table~\ref{tab:code-transfer}). On the untuned generator, whose decodes neither verifier saw in training, the same verifiers reach 0.62--0.78. In guided decoding on MBPP, the MBPP verifier is ahead of the GSM8K verifier for every generator, by 7.4 points for untuned Base, 9.4 for the SFT generator, 2.8 for Instruct and 5.2 for Dream (McNemar $p<10^{-5}$, $p<10^{-8}$, $p=0.092$ and $p=0.003$). On HumanEval, which contributes no training data, the differences are smaller, $+1.8$ points for untuned Base, $-0.6$ for the SFT generator, $+4.3$ for Instruct ($p=0.039$) and $+3.7$ for Dream ($p=0.33$). On the SFT generator's HumanEval decodes the code verifier scores the first block below $\theta$ for 98\% of the eventually wrong solutions and 58\% of the eventually right ones (87\% and 21\% on MBPP), and 73 of the 148 first regeneration events accept an alternative, so most of the budget is spent without changing the canvas.

\section{Calibration and Paired Intervention Analysis}
\label{sec:calibration}
We evaluate the probability of final-answer success under the continuation policy that generated each trajectory. Every observed block boundary inherits that trajectory's outcome. Sampled-policy evaluation uses 20,744 states from four responses per GSM8K test problem (5,276 responses in total, temperature 0.7). Greedy-policy evaluation uses 1,209 states from a 300-problem test subset. Commitment confidence, prompted Yes/No, and DiVR are evaluated on the same greedy states with the same outcome labels.

For each policy and signal separately, we fit positive-slope affine-logit Platt scaling \citep{platt1999probabilistic} by unweighted NLL on development states, with 7,671 states for the sampled-policy verifier and 957 for each greedy-policy signal. Fitted parameters are frozen for test evaluation. ECE uses ten equal-width bins, and Brier score and NLL retain the natural class ratio. We report state-weighted metrics, so longer trajectories contribute more observations. Final states form a separate group, and block-index groups contain only non-final states. No terminated trajectory is padded into later groups, unlike the pair-accuracy analysis in Section~\ref{sec:analysis}.

\begin{table}[t]
\caption{\textbf{Outcome prediction and post-hoc calibration.} GSM8K test states at observed block boundaries. Each signal has its own Platt map fitted on development states. Calibration preserves the ordering, so raw and calibrated AUROC agree.}
\label{tab:calibration}
\centering
\small
\setlength{\tabcolsep}{5pt}
\renewcommand{\arraystretch}{1.08}
\begin{tabular*}{\textwidth}{@{\extracolsep{\fill}}llrrrr@{}}
\toprule
State source / signal & Calibration & AUROC $\uparrow$ & ECE $\downarrow$ & Brier $\downarrow$ & NLL $\downarrow$ \\
\midrule
\multicolumn{6}{@{}l}{\textit{Sampled policy (20,744 states, 1,319 problems)}} \\
DiVR & Raw & 0.913 & 0.095 & 0.130 & 0.464 \\
 & Platt & 0.913 & 0.021 & 0.111 & 0.359 \\
\midrule
\multicolumn{6}{@{}l}{\textit{Greedy policy (1,209 states, 300 problems)}} \\
Commit confidence & Raw & 0.625 & 0.186 & 0.253 & 0.736 \\
 & Platt & 0.625 & 0.033 & 0.219 & 0.628 \\
Prompted Yes/No & Raw & 0.589 & 0.288 & 0.305 & 1.068 \\
 & Platt & 0.589 & 0.079 & 0.229 & 0.652 \\
DiVR & Raw & 0.909 & 0.093 & 0.125 & 0.447 \\
 & Platt & 0.909 & 0.027 & 0.110 & 0.358 \\
\bottomrule
\end{tabular*}
\end{table}

\begin{figure}[t]
    \centering
    \includegraphics[width=0.7\textwidth]{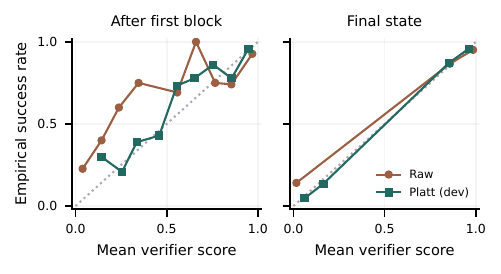}
    \caption{\textbf{Verifier reliability on greedy GSM8K states.} Mean score versus empirical success at the first non-final block and the final state (300 observations each), and markers are bins with at least ten observations. Platt parameters are fitted on development states.}
    \label{fig:calibration}
\end{figure}

On sampled-policy states, DiVR's ECE decreases from 0.095 to 0.021. The corresponding 95\% percentile intervals are [0.084, 0.106] and [0.014, 0.033], from 500 bootstrap resamples at the problem level, conditional on the fitted calibrator. After the first block, ECE falls from 0.116 to 0.031, and at completion from 0.071 to 0.019. On greedy states, first-block ECE falls from 0.136 to 0.067 and final-state ECE from 0.071 to 0.032 (Figure~\ref{fig:calibration}). A single global map does not guarantee calibration at each block. Calibrated confidence and Yes/No retain ECE 0.226 and 0.265 on late non-final states despite smaller pooled errors. Calibration and discrimination therefore need separate evaluation.

Positive-slope rescaling preserves candidate ordering. It changes a threshold's numeric interpretation, not which candidate is ranked highest. The main experiments retain the original scores and threshold, and no accuracy improvement is attributed to post-hoc calibration. This analysis does not establish calibration after adaptive regeneration or across benchmarks.

For the paired intervention study, we take a 150-problem test subset and save 436 greedy states from the first three blocks. At each state, we compare its original greedy outcome with one regeneration event, namely four alternatives at temperature 0.7, selection against the original by raw verifier score, and then greedy continuation. Each state is tested independently, rather than following the sequential three-event policy. Below score 0.1, success rises from 13.9\% to 26.1\% on 115 states, with 16 fixes and two regressions. At scores of at least 0.9, it rises from 92.4\% to 93.4\% on 211 states, with five fixes and three regressions. The threshold 0.5 selects 33.7\% of states, whose average gain is 9.5 points, versus 1.7 points for the remaining states. Each state uses one sampled alternative set. Problem-level bootstrap intervals (1,000 resamples) put the gain at threshold 0.5 at 9.5 points [4.2, 15.3].

\paragraph{Decision calibration.}
\citet{zhao2021calibrating} call predictions decision-calibrated for a family of decision rules if, for every rule, the loss the predictions anticipate for the chosen action equals the realized loss. Their predictions concern an outcome that the decision does not change. Here each action leads to a different state, so we predict the outcome of each action separately and require the prediction for the chosen action to match its realized outcome. We consider rules that regenerate when the raw score is below $\theta$.
\begin{itemize}
\item \textbf{Predicted success of an action.} We apply the greedy-policy Platt map, fitted on development states (Table~\ref{tab:calibration}), to the raw score of the state the action leads to, which is the current block when continuing and the retained block when regenerating.
\item \textbf{Realized success.} The outcome of the greedy continuation after that action.
\end{itemize}
Table~\ref{tab:decision-calibration} compares the two on the 436 paired states.
\begin{itemize}
\item \textbf{Chosen action.} Predicted and realized success differ by at most 2.4 points at every threshold.
\item \textbf{Regeneration gain.} The predicted gain matches the realized gain. At $\theta=0.5$ the difference is 0.8 points, with interval $[-4.1, 5.8]$.
\item \textbf{Selection.} Choosing the best of five candidates by score adds no optimism beyond a shared 2-point offset. After regenerating everywhere, predicted success is 72.1\% versus 70.2\% realized, and when continuing everywhere it is 67.8\% versus 65.8\%. The offset reflects the development-fitted map on this test subset, not selection.
\end{itemize}
All intervals are 95\% problem-level bootstrap intervals. The test covers one regeneration event per state, with candidates selected by raw score. It does not cover the sequential three-event policy or per-state repair probabilities.

\begin{table}[t]
\caption{\textbf{Decision calibration for regenerate-or-continue.} 436 paired greedy states from 150 GSM8K test problems. A rule regenerates when the raw score is below $\theta$, and predictions apply the development-fitted Platt map to the state each action leads to. The gap is predicted minus realized success of the chosen action, with a 95\% bootstrap interval.}
\label{tab:decision-calibration}
\centering
\small
\setlength{\tabcolsep}{7pt}
\begin{tabular}{@{}rrcc@{}}
\toprule
$\theta$ & Regen. & Gain (pred.\ / real.) & Chosen-action gap \\
\midrule
0.1 & 26\% & 11.0 / 12.2 & $+1.7$ [$-3.4$, $+7.0$] \\
0.3 & 30\% & 10.9 / 10.6 & $+2.0$ [$-2.6$, $+7.0$] \\
0.5 & 34\% & 10.3 / \phantom{0}9.5 & $+2.2$ [$-2.4$, $+7.8$] \\
0.7 & 38\% & \phantom{0}9.6 / \phantom{0}8.4 & $+2.4$ [$-2.4$, $+7.3$] \\
0.9 & 52\% & \phantom{0}7.8 / \phantom{0}7.6 & $+2.1$ [$-2.5$, $+6.8$] \\
\bottomrule
\end{tabular}
\end{table}

\clearpage
\section{Qualitative Examples}
\label{sec:examples}
\begin{table}[t]
\caption{\textbf{Regenerations that fix greedy errors.} GSM8K test problems on which the SFT generator's greedy answer is wrong and one regeneration corrects it. Each panel shows the block that fell below $\theta$ (Original), the four alternatives decoded from the same state, and their scores. The highest-scoring candidate (bold) is kept. $\hookleftarrow$ marks a line break.}
\label{tab:examples}
\centering
\scriptsize
\setlength{\tabcolsep}{4pt}
\renewcommand{\arraystretch}{1.12}
\begin{tabular}{@{}l r p{0.79\textwidth}@{}}
\toprule
Candidate & Score & Decoded block \\
\midrule
\multicolumn{3}{@{}p{0.985\textwidth}@{}}{\textbf{(a) Regeneration of the first block.} Jillian's handbag cost \$20 less than 3 times as much as her shoes cost. If her shoes cost \$80, how much did her bag cost? \emph{(Answer 220, greedy answer 200.)}} \\[1pt]
Original & 0.019 & \texttt{Three times the cost of her shoes is 3*\$80 =220 $\hookleftarrow$ \$20 less than 220 is 22} \\
Resample 1 & 0.395 & \texttt{Three times as much as her shoes cost is 3*\$80=\$240 $\hookleftarrow$ \$20 less than \$80 is \$84} \\
\textbf{Resample 2} & \textbf{0.990} & \texttt{Let x be the cost of the bag. $\hookleftarrow$ x = 3 * 80 = 240. $\hookleftarrow$ x = 240} \\
Resample 3 & 0.018 & \texttt{Three times the cost of her shoes is 3*\$80 = \$20 $\hookleftarrow$ \$20 less than \$20 is \$20-\$} \\
Resample 4 & 0.016 & \texttt{Three times the cost of her shoes is 3 * \$80 = \$200 $\hookleftarrow$ Jillian's handbag cost \$200 - \$} \\
\multicolumn{3}{@{}p{0.985\textwidth}@{}}{\textit{Final DiVR reasoning.} \texttt{Let x be the cost of the bag. $\hookleftarrow$ x = 3 * 80 = 240. $\hookleftarrow$ x = 240 - 20 = 220.} $\rightarrow$ answer 220 (correct).} \\
\midrule
\multicolumn{3}{@{}p{0.985\textwidth}@{}}{\textbf{(b) Regeneration of the second block, with the first kept.} Mulan has \$40. Her father gave her \$100. She buys two pairs of jeans at \$30 each and a bag for \$20. How much money does Mulan have left? \emph{(Answer 60, greedy answer 220.)}} \\[1pt]
Kept & 0.985 & \texttt{Mulan has \$40 + \$100 = \$140. $\hookleftarrow$ She spends \$30 x 2 = \$60 on} \\
Original & 0.039 & \texttt{jeans. $\hookleftarrow$ She has a total of \$140 + \$60 + \$20 = \$220. $\hookleftarrow$ </reasoning>} \\
\textbf{Resample 1} & \textbf{0.997} & \texttt{jeans. $\hookleftarrow$ She spends \$60 + \$20 = \$80 in total. $\hookleftarrow$ Mulan has \$140 - \$80 =} \\
Resample 2 & 0.035 & \texttt{jeans. $\hookleftarrow$ She has a total of \$140 + \$60 + \$20 = \$220. $\hookleftarrow$ </reasoning>} \\
Resample 3 & 0.035 & \texttt{jeans. $\hookleftarrow$ She has a total of \$140 + \$60 + \$20 = \$220. $\hookleftarrow$ </reasoning>} \\
Resample 4 & 0.997 & \texttt{jeans. $\hookleftarrow$ She spends \$60 + \$20 = \$80 in total. $\hookleftarrow$ Mulan has \$140 - \$80 =} \\
\multicolumn{3}{@{}p{0.985\textwidth}@{}}{\textit{Final DiVR reasoning.} \texttt{Mulan has \$40 + \$100 = \$140. $\hookleftarrow$ She spends \$30 x 2 = \$60 on jeans. $\hookleftarrow$ She spends \$60 + \$20 = \$80 in total. $\hookleftarrow$ Mulan has \$140 - \$80 = \$60 left.} $\rightarrow$ answer 60 (correct).} \\
\midrule
\multicolumn{3}{@{}p{0.985\textwidth}@{}}{\textbf{(c) Regeneration of the second block, with the first kept.} Eight more than four times the number of coffee mugs in the teacher's lounge is 40. How many coffee mugs are in the teacher's lounge? \emph{(Answer 8, greedy answer 0.)}} \\[1pt]
Kept & 0.966 & \texttt{Let x be the number of coffee mugs in the teacher's lounge. $\hookleftarrow$ Eight more than four times the number of coffee mugs in the teacher's lounge is} \\
Original & 0.034 & \texttt{40+8x. $\hookleftarrow$ Eight more than four times the number of coffee mugs in the teacher's lounge is 40, or 40} \\
Resample 1 & 0.984 & \texttt{4x + 8. $\hookleftarrow$ We know that 8 more than four times the number of coffee mugs in the teacher's lounge is 40,} \\
\textbf{Resample 2} & \textbf{0.997} & \texttt{4x+8. $\hookleftarrow$ We know that 4x+8=40. $\hookleftarrow$ Subtracting 8 from both sides, we get 4} \\
Resample 3 & 0.029 & \texttt{40+8x. $\hookleftarrow$ Thus four times the number of coffee mugs in the teacher's lounge is=40+x $\hookleftarrow$ 8x=40} \\
Resample 4 & 0.997 & \texttt{4x+8. $\hookleftarrow$ We know that 4x+8=40. $\hookleftarrow$ Subtracting 8 from both sides, we get 4} \\
\multicolumn{3}{@{}p{0.985\textwidth}@{}}{\textit{Final DiVR reasoning.} \texttt{Let x be the number of coffee mugs in the teacher's lounge. $\hookleftarrow$ Eight more than four times the number of coffee mugs in the teacher's lounge is 4x+8. $\hookleftarrow$ We know that 4x+8=40. $\hookleftarrow$ Subtracting 8 from both sides, we get 4x=32. $\hookleftarrow$ Dividing both sides by 4, we get x=8.} $\rightarrow$ answer 8 (correct).} \\
\bottomrule
\end{tabular}
\end{table}
Table~\ref{tab:examples} shows three GSM8K test problems on which a single regeneration corrects the SFT generator's greedy answer. The alternatives are reproduced from the recorded random state of each run.

\section{Prompt Templates}
\label{sec:prompts}
Every prompt is a single user turn in the generator's own chat template, with the generation prompt appended. A fixed response prefix follows it and belongs to the prompt, not to the decoded canvas. Line breaks are as used, and long lines are wrapped for display ($\hookrightarrow$). Braces mark inserted fields. LLaDA-8B-Base, LLaDA-8B-Instruct and their SFT adapters share one template. Dream-v0-Instruct-7B uses its ChatML template with the default system message, as listed below.
\begin{lstlisting}[style=prompt]
LLaDA: <|startoftext|><|start_header_id|>user<|end_header_id|>\n\n{content}<|eot_id|><|start_header_id|>assistant<|end_header_id|>\n\n{prefix}
Dream: <|im_start|>system\nYou are a helpful assistant.<|im_end|>\n<|im_start|>user\n{content}<|im_end|>\n<|im_start|>assistant\n{prefix}
\end{lstlisting}
The verifier adds no prompt of its own. It reads the same prompt followed by the current canvas. SR-SMC also uses only the generation prompt.

\paragraph{GSM8K and MATH-500.} The d1 instruction \citep{zhao2025d1} precedes the question, and the prefix is \texttt{<reasoning>}. The same content is used for the verifiers' training decodes.
\begin{lstlisting}[style=prompt]
You are a math expert. You will be given a question to solve. Solve it step by step. Wrap the final answer in a \boxed{}.
Respond in the following format:
<reasoning>
Your reasoning here
</reasoning>
<answer>
\boxed{...}
</answer>

{question}
\end{lstlisting}

\paragraph{Countdown.} The d1 Countdown instruction, verbatim, including its missing space before ``Respond''. The question lists the three numbers and the target. The prefix is \texttt{<reasoning>}.
\begin{lstlisting}[style=prompt]
Using only the provided numbers, create an arithmetic expression that evaluates to exactly the provided target number. You may use the operations +, -, *, and / as needed, but each number must be used exactly once. Think step-by-step. After reasoning, provide only your final expression inside \boxed{} tags without including an equals sign or the target number. For example: \boxed{a + b * c}Respond in the following format:
<reasoning>
Your reasoning here
</reasoning>
<answer>
\boxed{...}
</answer>

Numbers: [{n1}, {n2}, {n3}]
Target: {target}
\end{lstlisting}

\paragraph{HumanEval and MBPP.} The prefix opens a Python code fence followed by a newline. The program is the response up to the closing fence. HumanEval appends its prompt field, the function signature and docstring. MBPP lists the task text and its test assertions.
\begin{lstlisting}[style=prompt]
Complete the following Python function. Write the complete function, including its signature, inside a single ```python code block.

{HumanEval prompt}
\end{lstlisting}
\begin{lstlisting}[style=prompt]
You are an expert Python programmer. Write a Python function for the task below. Your code must pass the given tests. Put the complete code inside a single ```python code block.

Task: {text}
Tests:
{test_list, one assertion per line}
\end{lstlisting}

\paragraph{Sudoku diagnosis.} The d1 $4{\times}4$ instruction, verbatim, which begins with a newline. The trace SFT used the same prompt. The prefix is \texttt{<reasoning>}.
\begin{lstlisting}[style=prompt]
Please solve the following 4x4 Sudoku puzzle. The puzzle is provided as a 16-character string reading left-to-right, top-to-bottom, where '0' represents empty cells.

Rules:
- Fill empty cells with digits 1-4
- Each row must contain digits 1-4 exactly once
- Each column must contain digits 1-4 exactly once
- Each 2x2 box must contain digits 1-4 exactly once

Important: Your solution must be a COMPLETE 16-character string with only the digits 1-4, representing your final solved grid.

Respond in this exact format:
<reasoning>
Your step-by-step solving process
</reasoning>
<answer>
[16-character solution string with no spaces or separators]
</answer>


Solve the following Sudoku puzzle: {puzzle}
\end{lstlisting}

\paragraph{Prompted Yes/No and Prism's verifier.} Both use the generator itself as judge.
\begin{itemize}
\item \textbf{Fields.} The judge prompt is filled with the question and the solution decoded so far. A solution longer than 2,000 characters keeps its first 500 and last 1,500 characters around a \texttt{...\allowbreak[truncated]\allowbreak...} marker.
\item \textbf{Readout.} The prompt is wrapped in the chat template and one mask token is appended. The score is the two-way softmax of the largest Yes-variant logit (\texttt{Yes}, \texttt{\ Yes}, \texttt{YES}) against the largest No-variant logit, both read at that mask position.
\item \textbf{Prompts.} Mathematical tasks use Prism's judgment prompt. Code tasks use our shorter reviewer adaptation, the second listing.
\end{itemize}
\begin{lstlisting}[style=prompt]
You are an expert mathematician and competition judge. Your task is to evaluate a proposed mathematical solution for a given problem based on its logical rigor and accuracy.

[Math Problem]
{prompt}
[/Math Problem]

[Proposed Mathematical Solution]
{solution}
[/Proposed Mathematical Solution]

**Analysis Steps:**
1. Reasoning Validity: Are the logical steps and mathematical properties applied correctly?
2. Calculation Accuracy: Are the intermediate calculations or algebraic manipulations accurate?
3. Goal Alignment: Does the current reasoning path directly lead toward the final answer required by the problem?

**Conclusion**: Based on your analysis, is this solution path sound and likely to result in the correct final answer? Answer with a single word: Yes or No.
**Answer:**
\end{lstlisting}
\begin{lstlisting}[style=prompt]
You are an expert code reviewer. Your task is to evaluate whether a proposed Python solution correctly solves the task.

[Task]
{prompt}
[/Task]

[Proposed Solution]
{solution}
[/Proposed Solution]

**Conclusion**: Based on your analysis, is the solution likely to be fully correct? Answer with a single word: Yes or No.
**Answer:**
\end{lstlisting}
\end{document}